\pdfoutput=1
\documentclass[11pt]{article}

\usepackage[final]{acl}

\usepackage{times}
\usepackage{latexsym}

\usepackage{graphicx}  % for \resizebox
\usepackage{bm}        % for \bm
\usepackage{booktabs}  % for \toprule, \midrule, \cmidrule, \bottomrule

\usepackage[T1]{fontenc}
\usepackage[utf8]{inputenc}

\usepackage{microtype}

\usepackage{inconsolata}

\usepackage{graphicx}

\usepackage{booktabs}
\usepackage{adjustbox}
\usepackage{multirow}
\usepackage{amsmath}
\usepackage{amssymb}

\usepackage{cleveref}
\usepackage{algorithm}
\usepackage{algpseudocode}
\usepackage{xargs}
\usepackage{makecell}
\usepackage{enumitem}
\usepackage{longtable}
\usepackage{tcolorbox}
\usepackage{multicol}
\usepackage{setspace}
\usepackage{makecell}

\usepackage{pgfplots}
\usepackage{pgfplotstable}
\usepackage{caption}
\usepackage{subcaption}
\pgfplotsset{compat=1.17}

\title{CopySpec: Accelerating LLMs with Speculative Copy-and-Paste}

\author{Razvan-Gabriel Dumitru \\
  University of Arizona \\
  \texttt{razvandumm@gmail.com} \\\And
  Minglai Yang \\
  University of Arizona \\\And
  Vikas Yadav \\
  ServiceNow Research \\\And
  Mihai Surdeanu \\
  University of Arizona \\
}

\begin{document}
\maketitle
\begin{abstract}
We introduce \textit{CopySpec}, a simple yet effective technique to tackle the inefficiencies LLMs face when generating responses that closely resemble previous outputs or responses that can be verbatim extracted from context. 
\textit{CopySpec} identifies repeated sequences in the model’s chat history or context and speculates that the same tokens will follow, enabling seamless copying without compromising output quality and without requiring additional GPU memory. 
To evaluate the effectiveness of our approach, we conducted experiments using seven LLMs and five datasets: MT-Bench, CNN/DM, GSM8K, HumanEval, and our newly created dataset, \textit{MT-Redundant}. \textit{MT-Redundant}, introduced in this paper, transforms the second turn of MT-Bench into a request for variations of the first turn’s answer, simulating real-world scenarios where users request modifications to prior responses.
Our results demonstrate significant speed-ups: up to 2.35× on CNN/DM, 3.08× on the second turn of select \textit{MT-Redundant} categories, and 2.66× on the third turn of GSM8K’s self-correction tasks. Importantly, we show that \textit{CopySpec} integrates seamlessly with speculative decoding, yielding an average 49\% additional speed-up {\em over} speculative decoding for the second turn of \textit{MT-Redundant} across all eight categories.
While LLMs, even with speculative decoding, suffer from slower inference as context size grows, CopySpec leverages larger contexts to accelerate inference, making it a faster complementary solution. Our code and dataset are publicly available at https://github.com/RazvanDu/CopySpec.
\end{abstract}

\section{Introduction}

\begin{figure}[ht]
\vskip 0.2in
\begin{center}
\centerline{\includegraphics[width=\columnwidth]{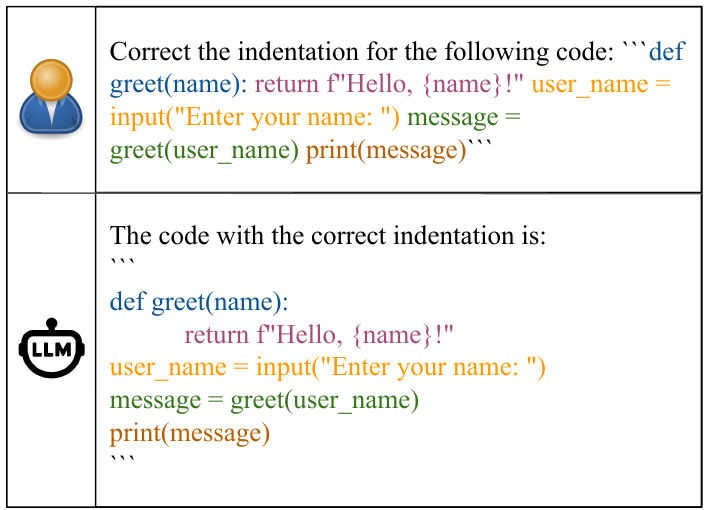}}
\caption{An example of redundant information, represented by blocks of the same color, that can be directly copied during inference without re-computation. This highlights the potential of our approach to make inference more efficient by leveraging repeated information, reducing computational overhead, and improving speed.}
\label{fig:redundant_example}
\end{center}
\vskip -0.4in
\end{figure}

\begin{figure*}[ht]
\vskip 0.2in
\begin{center}
\centerline{\includegraphics[width=\textwidth]{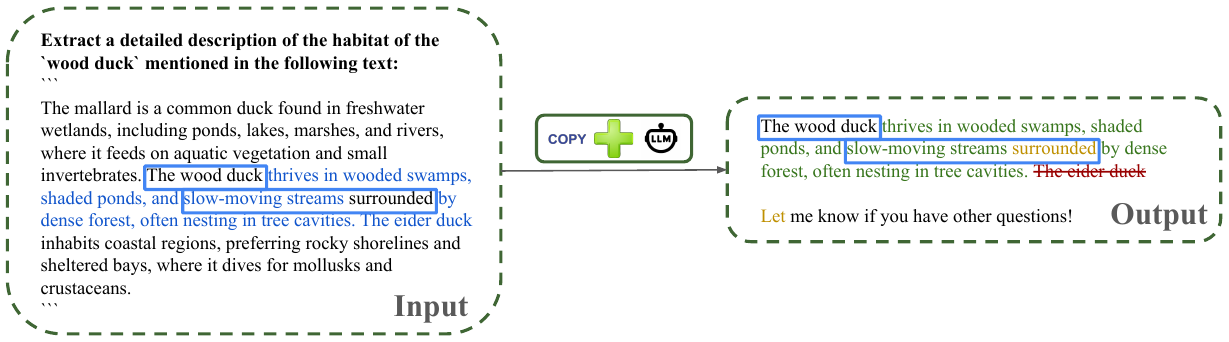}}
\caption{The figure illustrates the speculative copying process, \textit{CopySpec} applied to extract the habitat description of the "wood duck.". The input text provides the context and instructions. During generation, the system identifies sequences of 3 consecutive tokens (we use words as tokens here for illustrative simplicity) that repeat within the input. The \textcolor{blue}{blue} rectangle in the input highlights the matching token sequence detected, which serves as the starting point for speculative copying. From this match, the next 10 tokens are copied into the output.
In the output, the copied tokens are shown in \textcolor{blue}{blue} and validated through speculative copying. Tokens accepted by the model are highlighted in \textcolor[rgb]{0,0.87,0}{green}, continuing the description seamlessly, while rejected tokens are shown in \textcolor{red}{red} with a strikethrough. Extra tokens generated during the validation process are marked in \textcolor{orange}{yellow/gold}, demonstrating how the model extends the copied content as needed. This figure demonstrates how \textit{CopySpec} efficiently leverages repeated sequences to enhance text generation accuracy and speed by integrating both copied and dynamically generated content.}
\label{fig:copying_algorithm}
\end{center}
\vskip -0.2in
\end{figure*}

Large Language Models (LLMs) have revolutionized natural language processing (NLP), enabling great performance across a range of applications, including code generation, machine translation, and question answering. However, the computational demands of LLMs, particularly during inference, pose significant challenges for real-time applications and scalability in resource-constrained environments. Sequential token generation, a core bottleneck in standard decoding, limits throughput and increases latency. Speculative Decoding \cite{leviathan2023speculative, chen2023parallel} has emerged as a promising approach to mitigate this issue by employing a smaller draft model to generate multiple token sequences, which are then verified by the larger target model. Despite its potential, existing speculative decoding methods often fail to fully exploit the inherent redundancies in previously LLM-generated outputs and context. 
For example, in tasks such as summarization, retrieval augmented generation (RAG), and multi-turn conversations, etc., response generations include several text pieces verbatim from the context. In tasks such as code fixing, error can be just from a specific line of code which requires editing and regeneration while large portion of the code can be copy-pasted as it is. Further, only small edits are required in previous generation for frequent queries such as format changes, persona or style transfer, etc and in case of improving LLM outputs such as self-verification.  % Further, speculative decoding also require extra GPU memory to host a draft model or modifications to the original LLM. % In contrast, our approach, CopySpec, is simple, avoids the need for a draft model, and eliminates additional GPU overhead by directly copying tokens from the context or previous generations. 

In this work, we present \textit{CopySpec}, a mixed speculative decoding framework designed to exploit redundancies or repeated generations from previous iterations of responses or verbatim text available in context. \textit{CopySpec} incorporates a copying mechanism into the draft process, enabling the model to detect and exploit predictable patterns in token sequences (see Figure~\ref{fig:copying_algorithm} for a summary of the approach). By inferring subsequent tokens directly from prior context, \textit{CopySpec} reduces the computational burden by simply copy pasting the repeating response text instead of generating it. Importantly, \textit{CopySpec} can be combined with any speculative decoding method to further improve LLM inference speed. Basically, \textit{CopySpec} complements any existing speculative decoding method by simply copy pasting text where repeated text can be copied from context,  or else response is generated by the draft model. % Additionally, \textit{CopySpec} enhances the sampling-verification process with minimal computational overhead.
Our experiments on various benchmarks—including HumanEval \cite{chen2021evaluatinglargelanguagemodels}, CNN/DM \cite{DBLP:journals/corr/SeeLM17}, GSM8K \cite{cobbe2021gsm8k}, and MT Bench \cite{zheng2023judgingllmasajudgemtbenchchatbot}—demonstrate that CopySpec delivers up to an additional 49\% speed-up over speculative decoding, without compromising output quality. % This broad performance boost highlights its strong potential for real-world deployments. 

% \textbf{Key Contributions:}  
% \textbf{1)} \textit{CopySpec} introduces a novel framework that dynamically identifies and copies repeated token patterns, seamlessly integrating with any speculative decoding to improve inference efficiency. By leveraging a rolling hash mechanism, it efficiently speculates on larger token blocks with minimal computational overhead.  
% \textbf{2)} Our method achieves significant speedups with minimal overhead, requiring no changes to the LLM architecture or additional GPU memory, making it lightweight and practical for real-world use.  
% \textbf{3)} Evaluations across five datasets, including MT-Bench, CNN/DM, GSM8K, HumanEval, and MT-Redundant, demonstrate \textit{CopySpec}'s ability to deliver up to a 3.08× speedup in specific MT-Redundant categories and a 49\% speed-up on top of speculative decoding, without compromising output quality.

\section{Related Work}

\subsection{Complementing Speculative Decoding with Copying Mechanisms}

Recent work has advanced speculative decoding through multi-token verification \cite{leviathan2023speculative,cai2024medusa}, adaptive pipelines \cite{liu2024adaptive}, token tree structures \cite{specinfer2023}, pipelined execution \cite{predictive2023pipelined}, draft-target distillation \cite{distillspec2023}, and retrieval-based validation \cite{rest2023retrievalaugmented}, with further gains from optimal transport \cite{sun2024spechubprovableaccelerationmultidraft} and early-layer reuse \cite{he2023speed}.

Complementary efforts accelerate decoding via prompt reuse: PLD and PLD$+$ copy prompt $n$-grams \cite{saxena2023prompt,somasundaram2024pldacceleratingllminference}, PPD overlaps streams for GPU efficiency \cite{chen2024hardwareawareparallelpromptdecoding}, look-ahead decoding verifies disjoint blocks \cite{fu2024break}, and token recycling repurposes failed tokens \cite{luo2024turningtrashtreasureaccelerating}. The closest to our method is speculative decoding with suffix alignment \cite{hu2024samdecodingspeculativedecoding}, though it introduces significantly more algorithmic complexity and overhead.

\emph{CopySpec} complements all of the above with minimal overhead. Instead of restricting reuse to prompt-boundary $n$-grams, it uses a rolling hash to find any repeated $\gamma$-token suffix in context and speculates on the next block. These candidate tokens are verified directly, enabling integration with any speculative decoding setup and yielding additional speedups without compromising output quality.

\subsection{Copying Mechanisms in Language Models}

Copying mechanisms are widely adopted in NLP to handle tasks that require replicating predictable patterns or segments. \citet{gu2016incorporating} introduced CopyNet, a method that enables RNN sequence-to-sequence models to predict words based on a mixed probabilistic model of two modes, where one selects words from the source sequence.
% selectively generate or copy tokens, addressing challenges like rare or repetitive sequences. 
Similarly, in summarization tasks, Pointer Networks \cite{vinyals2015pointer} and Pointer-Generator Networks \cite{DBLP:journals/corr/SeeLM17} demonstrated the effectiveness of combining copying and generation to improve output fidelity and handle out-of-vocabulary tokens. 

% More recently, \citet{mccoy2023how} analyzed the extent to which transformers copy from their training data, providing insights into copying behaviors in modern LLMs. \citet{jelassi2024repeat} showed that transformers outperform state space models in copying repetitive patterns. 

% Lastly, in a different domain, \citet{andronov2024acceleratinginferencestringgenerationbased} introduced a copying mechanism into a transformer-based encoder-decoder that models chemical reactions by observing that  portions of the input chemicals often remain unchanged in the output.

In line with motivations from previous approaches that have emphasized importance of copying mechanisms in various applications, our work provides the simplest and most effective method for LLM inference. \textit{CopySpec} integrates a 
% intelligent % ms: no need to call your ideas intelligent
copying mechanism into speculative decoding, effectively reducing redundancy and enhancing efficiency across a wide range of tasks. By leveraging repeated patterns in the model's context, \textit{CopySpec} introduces a novel approach to accelerate inference while maintaining high performance.

% ms: removed the text below. seems weak to me
% This approach is not an advancement of copying methods but a novel application that builds upon these foundational ideas.

\begin{table*}[!ht]
\centering
\footnotesize
\resizebox{\textwidth}{!}{%
\begin{tabular}{l l c c c c c c}
\toprule
\textbf{Model} & \textbf{Variant} & \textbf{Metrics}  & \textbf{MT-Redundant} & \textbf{CNN/DM} & \textbf{GSM8K} & \textbf{MT-Bench} & \textbf{HumanEval} \\
\textit{(Instruct)} &    &    & \textit{0-shot} & \textit{0-shot} & \textit{3-turn} & \textit{0-shot} & \textit{0-shot} \\
 &    &    & GPT-4 Score (↑)  & ROUGE-L (↑) & Accuracy (↑) & GPT-4 Score (↑) & Accuracy (↑) \\
\midrule
\multirow{4}{*}{\textbf{Qwen2.5-72B}} & Both & Score & 9.28 & 0.213 & 96\% & 9.18 & 87.8\% \\
\cmidrule(lr){2-8}
            & CopySpec & Tokens/Sec & \textbf{6.42}$\pm0.01$ & \textbf{8.68}$\pm0.01$ & \textbf{7.01}$\pm0.01$ & \textbf{5.55}$\pm0.01$ & \textbf{7.01}$\pm0.01$ \\
            &          & Copied  & 32.35\% & 82.48\% & 47.59\% & 20.53\% & 37.47\% \\
            %\cdashline{2-8}
            & Base model & Tokens/Sec & 4.82$\pm0.01$ & 3.70$\pm0.01$ & 4.55$\pm0.01$ & 4.83$\pm0.01$ & 4.98$\pm0.01$ \\
%\addlinespace
\midrule
\multirow{4}{*}{\textbf{Qwen2.5-32B}} & Both & Score & 9.10 & 0.214 & 93\% & 8.97 & 89.6\% \\
\cmidrule(lr){2-8}
            & CopySpec & Tokens/Sec & \textbf{13.82}$\pm0.01$ & \textbf{18.34}$\pm0.03$ & \textbf{14.84}$\pm0.01$ & \textbf{12.15}$\pm0.01$ & \textbf{14.41}$\pm0.01$ \\
            &          & Copied  & 33.17\% & 81.82\% & 44.93\% & 22.61\% & 34.23\% \\
            & Base model & Tokens/Sec & 10.26$\pm0.01$ & 7.79$\pm0.01$ & 9.76$\pm0.01$ & 10.29$\pm0.01$ & 10.46$\pm0.01$ \\
%\addlinespace
\midrule
\multirow{4}{*}{\textbf{Qwen2.5-7B}} & Both & Score & 8.53 & 0.230 & 85\% & 8.41 & 82.3\% \\
\cmidrule(lr){2-8}
            & CopySpec & Tokens/Sec & \textbf{54.05}$\pm0.11$ & \textbf{47.15}$\pm0.08$ & \textbf{63.37}$\pm0.54$ & \textbf{46.85}$\pm0.08$ & \textbf{48.79}$\pm0.01$ \\
            &          & Copied  & 34.42\% & 65.67\% & 53.01\% & 22.86\% & 32.68\% \\
            & Base model & Tokens/Sec & 39.88$\pm0.02$ & 25.25$\pm0.05$ & 38.58$\pm0.03$ & 39.98$\pm0.01$ & 33.63$\pm0.06$ \\
%\addlinespace
\midrule
\multirow{4}{*}{\textbf{Llama3.1-70B}} & Both & Score & 8.74 & 0.204 & 90\% & 8.72 & 77.4\% \\
\cmidrule(lr){2-8}
            & CopySpec & Tokens/Sec & \textbf{6.57}$\pm0.01$ & \textbf{5.49}$\pm0.01$ & \textbf{6.06}$\pm0.01$ & \textbf{5.83}$\pm0.01$ & \textbf{6.24}$\pm0.01$ \\
            &          & Copied  & 31.42\% & 38.35\% & 30.07\% & 21.83\% & 27.54\% \\
            & Base model & Tokens/Sec & 4.98$\pm0.01$ & 4.19$\pm0.01$ & 4.77$\pm0.01$ & 4.98$\pm0.01$ & 5.05$\pm0.01$ \\
%\addlinespace
\midrule
\multirow{4}{*}{\textbf{Llama3.1-8B}} & Both & Score & 8.03 & 0.185 & 79\% & 7.54 & 65.9\% \\
\cmidrule(lr){2-8}
            & CopySpec & Tokens/Sec & \textbf{49.28}$\pm0.08$ & \textbf{37.44}$\pm0.19$ & \textbf{49.60}$\pm0.01$ & \textbf{45.84}$\pm0.07$ & \textbf{46.49}$\pm0.48$ \\
            &          & Copied  & 35.45\% & 38.32\% & 38.01\% & 30.01\% & 26.44\% \\
            & Base model & Tokens/Sec & 35.51$\pm0.01$ & 26.57$\pm0.11$ & 35.19$\pm0.09$ & 35.43$\pm0.01$ & 37.57$\pm0.22$ \\
%\addlinespace
\midrule
\bottomrule
\end{tabular}
}
\caption{Performance comparison across five models (Qwen2.5-72B, Qwen2.5-32B, Qwen2.5-7B, Llama3.1-70B, and Llama3.1-8B) using CopySpec versus baseline configurations on multiple datasets, including MT-Redundant, CNN/DM, GSM8K, MT-Bench, and HumanEval. Metrics include model-specific scores 
(GPT-4, using the 0613 checkpoint: Score, ROUGE-L, Accuracy), token generation rates (tokens/sec), and percentage of tokens copied. Results demonstrate the effectiveness of CopySpec in enhancing computational efficiency without compromising quality, achieving notable speed-ups and high token-copying rates in diverse tasks and model sizes.}

%Performance comparison across five models (Qwen2.5-72B, Qwen2.5-32B, Qwen2.5-7B, Llama3.1-70B, and Llama3.1-8B) using CopySpec versus baseline configurations on multiple datasets, including MT-Redundant, CNN/DM, GSM8K, MT-Bench, and HumanEval. Metrics include model-specific scores 
%(GPT-4, using the 0613 checkpoint: Score, ROUGE-L, Accuracy), token generation rates (tokens/sec), and percentage of tokens copied. Notably, larger Qwen models produce more extractive summaries on CNN/DM, leading to slightly lower ROUGE-L scores. Results demonstrate the effectiveness of CopySpec in enhancing computational efficiency without compromising quality, achieving notable speed-ups and high token-copying rates in diverse tasks and model sizes.
\label{tab:performance}
\end{table*}

\section{Method}

Our method operates on the assumption that if the last $\gamma$ tokens generated by an LLM appear in the context, the tokens that followed them in the input context are likely to follow again in the output. Figures~\ref{fig:redundant_example} and~\ref{fig:copying_algorithm} illustrate this concept, % where colored blocks represent matching segments between the input and output. % ms: redundant with the captions
By accurately identifying the start of such a segment, we can simply copy-paste all tokens within the block in a single pass, bypassing the need for a draft model to produce them incrementally. In the following subsections, we detail the implementation of this approach and its integration into a Speculative Decoding framework, demonstrating how it achieves substantial speed-ups.

\subsection{Identifying the Tokens to Copy}

To efficiently detect when the model begins generating a block that has already been produced earlier, we maintain a hash map containing all subsequences of $\gamma$ tokens from the context. During the generation process, we search this hash map for matches to the last $\gamma$ tokens generated. Adding a new tuple of tokens to the hash map and searching for a match after each generated token has a time complexity of $O(\gamma)$. Since $\gamma$ is typically set to a small value (e.g., $3$ or $5$), the computational overhead for processing new tokens and finding matches is minimal and independent of the context size. This stands in contrast to alternative approaches that require searching the entire context for the last substring, which can become computationally expensive as the context grows.

Our technique efficiently leverages larger contexts, allowing inference to become faster as the context size increases. By keeping $\gamma$ fixed, we ensure a balance between efficiency and precision. Additionally, we explored methods to utilize partial outputs without revealing the complete results and investigated how the semantic relationship between the preceding $\gamma$ tokens and the subsequent token can guide the optimal choice of $\gamma$. Further details are provided in Appendix~\ref{sec:cs}.

\subsection{Speculating on the Matched Tokens}

After identifying a match of $\gamma$ tokens in the context, we extract the subsequent tokens from the context, as shown in Figure~\ref{fig:copying_algorithm}. These extracted tokens, which we call $S_{copyspec}$, essentially simulate the behavior of a draft model where the probability for each token in $S_{copyspec}$ is treated as 100\%. But instead of generating, \textit{CopySpec} pastes the output from context, thus avoiding the expensive computation of generation by draft model. \footnote{If multiple matches exist for the last $\gamma$ tokens, we simply select the first, though more efficient strategies may exist.}

$S_{copyspec}$ is then verified directly by the main LLM. Each verification yields $\tau$ tokens that align with the LLM’s ongoing generation, along with one additional guaranteed token. This approach mirrors vanilla speculative decoding \cite{leviathan2023speculative}, where speculative tokens are appended to the context, and the longest prefix matching the LLM’s output is accepted. In Figure~\ref{fig:copying_algorithm}, $S_{copyspec}$ is highlighted in blue. The output shows the $\tau$ accepted tokens in green, the extra guaranteed token in gold, and any rejected tokens in red.
% \textcolor{blue}{This process effectively treats the copied tokens as a ``perfect prediction,'' ensuring efficient token generation when patterns are detected.}

After each newly pasted or generated token, or copying attempt, we re-evaluate the last $\gamma$ tokens in the context to identify a new match, allowing the model to utilize longer copyable blocks whenever possible. This eliminates the need for manual token generation between copying steps.

If any tokens in $S_{copyspec}$ fail the verification step, the model generates a new token that diverges from the previously matched tokens. This ensures that the next copying attempt yields a different match, preventing the model from getting stuck in repetitive loops.

\subsection{Merging with Speculative Decoding}
\label{sec:mergingspecdec}

To further enhance our technique, we have integrated it within a vanilla Speculative Decoding framework. At each step of the generation process, we attempt to find matches in the context. If a match for the last $\gamma$ tokens is found, we use $S_{copyspec}$ as draft tokens, effectively simulating a draft model with perfect confidence in those tokens. If no match is identified, we rely on a smaller draft model to generate $\tau_2$ draft tokens. This dual approach allows us to dynamically choose between leveraging repetitive patterns and utilizing speculative decoding for efficient token generation in contexts with little or no redundancy.

This integration provides the best of both worlds: Speculative Decoding accelerates inference when the context size is small or lacks redundancy, while CopySpec builds on this speed-up in subsequent steps by taking advantage of repetitive patterns as the context size increases. 

It is also worth noting that when used as a stand-alone method, CopySpec does not require a draft model. This eliminates the need for additional GPU memory or modifications to the model, making it lightweight and easy to deploy. We also explore the interplay between these techniques in Section~\ref{sec:analyses}, while Section~\ref{sec:kv_caching_appendix} provides a detailed account of the full implementation, including key-value caching. Furthermore, we mainly focus on the integration with vanilla speculative decoding\cite{leviathan2023speculative} in the paper for generalizability but we also integrate it as part of the EAGLE\cite{li2025eaglespeculativesamplingrequires} framework and compare it against baselines in Appendix~\ref{sec:comp_eagle} to showcase the technique's potential.

\section{Experiments}

\subsection{Models and Hyperparameters}

We evaluated our technique on five instruction-tuned LLMs: Qwen2.5-{72B, 32B, 7B} \cite{qwen2025qwen25technicalreport}, LLaMa3.1-{70B, 8B} \cite{grattafiori2024llama3herdmodels}, Vicuna-v1.3-{13B, 7B}\cite{zheng2023judgingllmasajudgemtbenchchatbot}, using 4 A100 GPUs with a batch size of 1 and temperature 0. Unless stated, $\gamma=$3, $|S_{copyspec}|=$10, and max generation length to 1024.

\subsection{Evaluation Datasets}

We evaluated our technique on five datasets, each targeting specific aspects of model performance: MT-Redundant, CNN/DM, GSM8K, MT-Bench, and HumanEval. MT-Redundant was designed to emphasize prompts requiring small variations to previous outputs, while CNN/DM focuses on extractive summarization. GSM8K evaluates the model's self-correction capabilities, MT-Bench highlights scenarios with minimal copying potential to measure the technique's overhead, and HumanEval assesses coding capabilities. To accommodate the increased computational demands of GSM8K and CNN/DM and our limited GPU resources, we restricted these datasets to 100 samples, ensuring they were of comparable size to the other datasets. For HumanEval, we employed the same instruction format as presented in EvalPlus \cite{evalplus}. Detailed descriptions of all prompts used in our experiments are provided in Appendixes~\ref{sec:prompts} and~\ref{sec:redundant_dataset}.

\subsection{Synergy with External Drafters}  

A detailed head-to-head between CopySpec and the strongest publicly
available accelerators—EAGLE~\cite{li2025eaglespeculativesamplingrequires}, PLD~\cite{somasundaram2024pldacceleratingllminference} and SAM-D~\cite{hu2024samdecodingspeculativedecoding}—appears in
Appendix~\ref{sec:comp_eagle}.  The headline figures
(Table~\ref{tab:eagle_summary}) show that combining EAGLE with
span-level copying lifts Vicuna-7B throughput on MT-Redundant from
$82\!\rightarrow\!95$ TPS, a \textit{2.4$\times$} boost over the
greedy decoder.

\subsection{MT-Redundant}

Most existing NLP datasets focus on tasks involving either single-turn interactions or scenarios where the model must entirely change its response in the second turn. These setups fail to capture realistic use cases where a user might request slight variations or refinements to a previous answer. To address this gap and highlight the capabilities of our technique, we introduce a new dataset, \textit{MT-Redundant}. 

\textit{MT-Redundant} is derived by modifying the second turn of MT-Bench \cite{zheng2023judgingllmasajudgemtbenchchatbot}. In our dataset, the second turn replaces the original question with a prompt asking the model to review its previous answer and make specific adjustments or variations. This modification simulates real-world scenarios where incremental refinement or elaboration is required. Example prompts from the dataset are provided in Appendix~\ref{sec:redundant_dataset}. For questions with reference answers, we retained the original reference for the first turn and created a new reference answer for the second turn to align with the revised prompts.

Our dataset spans a diverse range of practical use cases, categorized into eight groups: Coding, Extraction, Humanities, Math, Reasoning, Roleplay, STEM, and Writing. These categories reflect realistic tasks encountered in various domains. Additionally, we adopted the same evaluation procedure from MT-Bench to ensure consistency and comparability of results.

By creating \textit{MT-Redundant}, we aim to bridge the gap between artificial benchmarks and practical applications, providing a more representative evaluation for techniques like CopySpec in multi-turn interactions with repetitive information.

\section{Discussion of Results}

We analyze our main results in Table~\ref{tab:performance}, which show the impact of our method on performance and the percentage of tokens copied across five LLMs and datasets. The results are aggregated for all turns in MT-Redundant and MT-Bench (two turns each) and the self-correction process in GSM8K (three turns). Speedups range from 1.15$\times$ on MT-Bench, which has minimal redundancy, using Qwen2.5-72B-Instruct, to 2.35× on CNN/DM.

\begin{table}[h]
\centering
%\footnotesize
\resizebox{\columnwidth}{!}{%
\begin{tabular}{lcccc}
\toprule
& \multicolumn{2}{c}{\textbf{Turn 1}} & \multicolumn{2}{c}{\textbf{Turn 2}} \\
\cmidrule(lr){2-3} \cmidrule(lr){4-5}
\textbf{Category} & \textbf{Base Model} & \textbf{CopySpec} & \textbf{Base Model} & \textbf{CopySpec} \\
\midrule
Coding & 5.12 $\pm 0.01$ & \textbf{5.62} $\pm 0.01$ & 4.61 $\pm 0.01$ & \textbf{9.33} $\pm 0.01$ \\
Extraction & 4.76 $\pm 0.01$ & \textbf{5.65} $\pm 0.01$ & 4.58 $\pm 0.01$ & \textbf{8.30} $\pm 0.01$ \\
Humanities & 5.09 $\pm 0.01$ & \textbf{5.33} $\pm 0.01$ & 4.55 $\pm 0.01$ & \textbf{5.45} $\pm 0.01$ \\
Math & 5.17 $\pm 0.01$ & \textbf{5.84} $\pm 0.01$ & 4.75 $\pm 0.01$ & \textbf{10.14} $\pm 0.01$ \\
Reasoning & 5.08 $\pm 0.01$ & \textbf{5.69} $\pm 0.01$ & 4.65 $\pm 0.01$ & \textbf{10.84} $\pm 0.01$ \\
Roleplay & 5.08 $\pm 0.01$ & \textbf{5.14} $\pm 0.01$ & 4.58 $\pm 0.01$ & \textbf{14.10} $\pm 0.03$ \\
Stem & 5.12 $\pm 0.01$ & \textbf{5.37} $\pm 0.01$ & 4.61 $\pm 0.01$ & \textbf{6.78} $\pm 0.01$ \\
Writing & 5.12 $\pm 0.01$ & \textbf{5.13} $\pm 0.01$ & 4.65 $\pm 0.01$ & \textbf{10.59} $\pm 0.01$ \\
\midrule
\textbf{Average} & 
5.07 $\pm 0.01$ & 
\textbf{5.47} $\pm 0.01$ & 
4.62 $\pm 0.01$ & 
\textbf{9.44} $\pm 0.01$ \\
\bottomrule
\end{tabular}
}
\caption{Comparison of model speeds measured in tokens/sec across two turns on MT-Redundant using CopySpec (Qwen2.5-72B-Chat, $\gamma = 3$). The technique leads to an overall speed-up of 2.04 on the second turn.}
\label{tab:category_qwen72B_redundant}
\end{table}

While these results are notable, the key strength of our approach is its ability to enhance performance as the context grows. To illustrate this, we look at per-turn performance and analyze the effect of varying hyperparameters on the technique's effectiveness in a wide range of use-cases.

\begin{table*}[h]
\centering
\resizebox{\textwidth}{!}{%
\begin{tabular}{lcccccccc}
\toprule
& \multicolumn{4}{c}{\textbf{Turn 1}} & \multicolumn{4}{c}{\textbf{Turn 2}} \\
\cmidrule(lr){2-5} \cmidrule(lr){6-9}
\textbf{Category} 
& \textbf{Base Model} 
& \textbf{Spec.\ Dec.} 
& \textbf{Spec.\ Dec.} 
& \textbf{Spec.\ Dec. } 
& \textbf{Base Model} 
& \textbf{Spec.\ Dec.} 
& \textbf{Spec.\ Dec.} 
& \textbf{Spec.\ Dec.} \\
& & & \textbf{+ Copy ($\gamma=3$)} & \textbf{+ Copy ($\gamma=5$)} & & & \textbf{+ Copy ($\gamma=3$)} & \textbf{+ Copy ($\gamma=5$)} \\
\midrule
Coding      & 10.87 $\pm 0.01$& 15.88 $\pm 0.01$& 15.85 $\pm 0.08$& \textbf{16.17} $\pm 0.01$ & 9.73 $\pm 0.01$& 14.74 $\pm 0.01$& 22.12 $\pm 0.03$& \textbf{22.17} $\pm 0.08$ \\
Extraction  & 10.09 $\pm 0.01$& 14.07 $\pm 0.02$& \textbf{15.49} $\pm 0.08$ & 15.41 $\pm 0.01$& 9.79 $\pm 0.01$& 14.50 $\pm 0.02$& 18.56 $\pm 0.10$& \textbf{18.69} $\pm 0.01$ \\
Humanities  & 10.85 $\pm 0.01$& 13.62 $\pm 0.03$& \textbf{13.86} $\pm 0.02$ & 13.88 $\pm 0.01$& 9.75 $\pm 0.01$& 12.79 $\pm 0.02$& \textbf{13.66} $\pm 0.02$ & 13.73 $\pm 0.03$\\
Math        & 11.01 $\pm 0.01$& 16.94 $\pm 0.05$& 17.23 $\pm 0.01$& \textbf{17.30} $\pm 0.02$ & 10.05 $\pm 0.01$& 15.45 $\pm 0.01$& \textbf{24.28} $\pm 0.03$ & 24.11 $\pm 0.04$\\
Reasoning   & 10.80 $\pm 0.02$& 13.96 $\pm 0.02$& 14.18 $\pm 0.20$& \textbf{14.24} $\pm 0.07$ & 10.05 $\pm 0.01$& 14.20 $\pm 0.01$& \textbf{21.56} $\pm 0.09$ & 20.35 $\pm 0.07$\\
Roleplay    & 10.90 $\pm 0.01$& 12.80 $\pm 0.04$& 12.84 $\pm 0.01$& \textbf{12.97} $\pm 0.01$ & 9.93 $\pm 0.01$& 15.14 $\pm 0.03$& \textbf{29.02} $\pm 0.01$ & 27.95 $\pm 0.09$\\
Stem        & 10.90 $\pm 0.01$& 14.25 $\pm 0.03$& 14.33 $\pm 0.01$& \textbf{14.56} $\pm 0.01$ & 9.83 $\pm 0.01$& 13.94 $\pm 0.01$& 17.22 $\pm 0.02$& \textbf{17.26} $\pm 0.02$ \\
Writing     & 10.92 $\pm 0.01$& 12.56 $\pm 0.05$& 12.64 $\pm 0.01$& \textbf{12.73} $\pm 0.01$ & 9.94 $\pm 0.01$& 14.96 $\pm 0.02$& \textbf{26.64 $\pm 0.04$} & 25.08 $\pm 0.08$\\
\midrule
\textbf{Average}
& 10.79 $\pm 0.01$& 14.26 $\pm 0.03$& 14.55 $\pm 0.05$& \textbf{14.66} $\pm 0.02$ & 9.88 $\pm 0.01$& 14.47 $\pm 0.02$& \textbf{21.63} $\pm 0.04$ & 21.17 $\pm 0.05$\\
\bottomrule
\end{tabular}%
}
\caption{Comparison of decoding strategies in MT-Redundant across two turns, using Qwen2.5-32B-Instruct as the target model and Qwen2.5-7B-Instruct as the draft model. The table demonstrates the impact of CopySpec integration at different parameter settings ($\gamma=3$ and $\gamma=5$), with the draft model generating 3 tokens. Results highlight significant improvements in speed and token copying efficiency, particularly in the second turn, due to the interplay between speculative copying and draft model generation.}
\label{tab:specdec_results}
\end{table*}

\subsection{Speed-up by Turn and Category}

We begin our analysis by examining the speedups achieved on MT-Redundant for both the first and second turns, as summarized in Table~\ref{tab:category_qwen72B_redundant}. The results indicate a substantial average speedup of 2.04\(\times\) for the second turn, compared to a more modest speedup of 1.08\(\times\) for the first turn. Notably, the performance in tokens per second (TPS) achieved by the model increases for the second turn, which features a larger context size. In contrast, the baseline model experiences a decline in TPS as the context size increases. Another notable aspect is that the observed speedup is highly dependent on the specific use case. For instance, we observe speedups as low as 1.2\(\times\) in the Humanities category and as high as 3.08\(\times\) for Roleplay. However, regardless of the use case, the speedup for the second turn remains consistently positive across all models for both MT-Redundant and MT-Bench.

The results for all five models on MT-Redundant and MT-Bench are detailed in Appendix~\ref{sec:extra_speedup_mtredundant} and~\ref{sec:extra_speedup_mtbench} respectively. On average, the second round of MT-Redundant achieves a significant 91\% speedup across all models, compared to 31\% for MT-Bench. Notably, even on MT-Bench, which has less redundancy, the TPS achieved by CopySpec in the second turn is almost always higher than the baseline model's TPS in the first turn.

\subsection{The Effect of Gamma (\texorpdfstring{$\gamma$}{gamma})}
\label{sec:effectgamma}

\begin{figure}[ht]
\begin{center}
\includegraphics[width=0.45\textwidth]{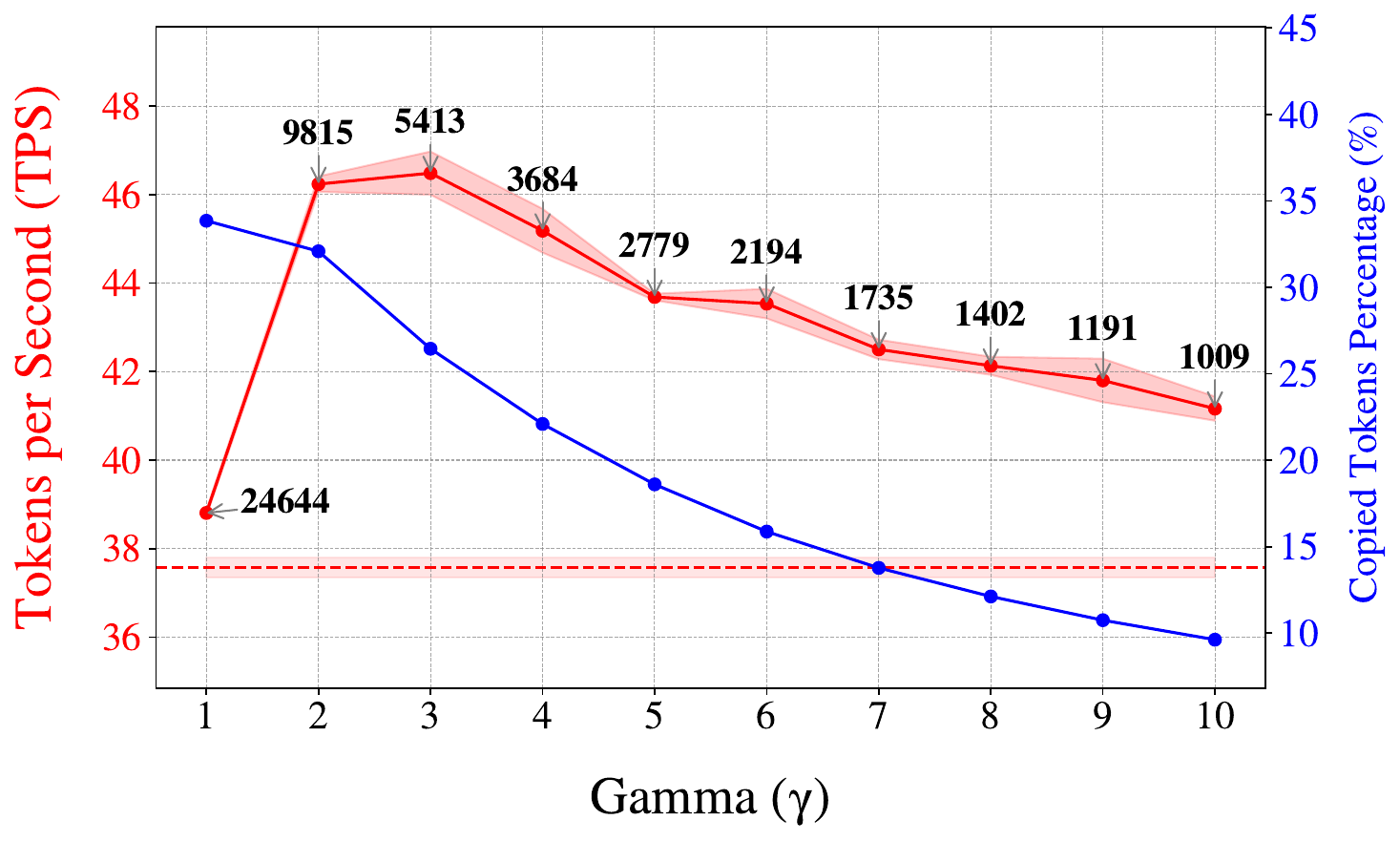}
\caption{This figure shows how the copying parameter $\gamma$ affects HumanEval performance using LLaMa3.1-8B-Instruct. The solid red line indicates tokens per second (TPS) with standard deviation shading; the dashed red line marks baseline TPS. The blue line shows the percentage of successfully copied tokens, with adjacent numbers indicating copying attempts.}
\label{fig:speedup_gamma_llama8B_humaneval}
\end{center}
\end{figure}

We begin our analysis with Figure~\ref{fig:speedup_gamma_llama8B_humaneval}, which illustrates the tokens per second as a red line, alongside the percentage of tokens copied out of the total tokens generated, represented by a blue line for the LLaMa3.1-8B model on HumanEval. The numbers adjacent to the dots indicate the number of attempts made to copy tokens. The figure demonstrates that as $\gamma$ increases, a smaller percentage of tokens is accepted, but the number of copying attempts decreases exponentially, leading to a significantly smaller overhead. A similar pattern is observed for MT-Redundant and MT-Bench, as presented in
%\cref{fig:speedup_gamma_llama8B_mtredundant} and 
Figure~\ref{fig:speedup_gamma_llama8B_mtbench} in the appendix.

%\vspace{-22mm}

Empirically, the optimal value of $\gamma$ across datasets is three, with two yielding similar performance. It is also worth noting that all $\gamma$ values ranging from 2 to 10, consistently results in significantly higher overall TPS, even across both turns on MT-Redundant and MT-Bench.

\subsection{Number of Tokens to Copy and Overhead}

We evaluate the impact of the number of tokens copied on performance and estimate CopySpec's overhead by setting the number of copied tokens to zero, isolating the cost of token searching. Results in Table~\ref{tab:tps_overhead} show minimal overhead  with differences from the base model nearly within the margin of error. Among the hyperparameters studied, setting $|S_{copyspec}=10|$ delivers the best performance, while larger values, such as 50 or 100, increase overhead and reduce tokens-per-second efficiency.

\begin{table}[H]
\centering
\resizebox{0.8\columnwidth}{!}{%
\footnotesize
\begin{tabular}{@{}ccc@{}}
\toprule
\textbf{Tokens Copied}    & \textbf{MT-Redundant}        & \textbf{MT-Bench}          \\ \midrule
Base Model            & 35.63 $\pm 0.04$            & 35.30 $\pm 0.16$           \\
0                         & 35.46 $\pm 0.01$            & 35.22 $\pm 0.04$           \\
\midrule
5                        & 47.64 $\pm 0.11$            & 44.69 $\pm 0.11$           \\
10                        & \textbf{49.52} $\pm 0.01$            & \textbf{45.74} $\pm 0.01$           \\
50                        & 45.56 $\pm 0.08$            & 41.59 $\pm 0.04$           \\
100                       & 39.41 $\pm 0.06$            & 35.76 $\pm 0.05$           \\ \bottomrule
\end{tabular}%
}
\caption{Tokens-per-second (TPS) on MT-Redundant and MT-Bench with LLaMa3.1-8B-Instruct. Results demonstrate that copying 10 tokens achieves optimal performance, while larger copying attempts introduce overhead, reducing overall efficiency.}
\label{tab:tps_overhead}
\end{table}

\begin{figure}[H]
\begin{center}
\centerline{\includegraphics[width=\columnwidth]{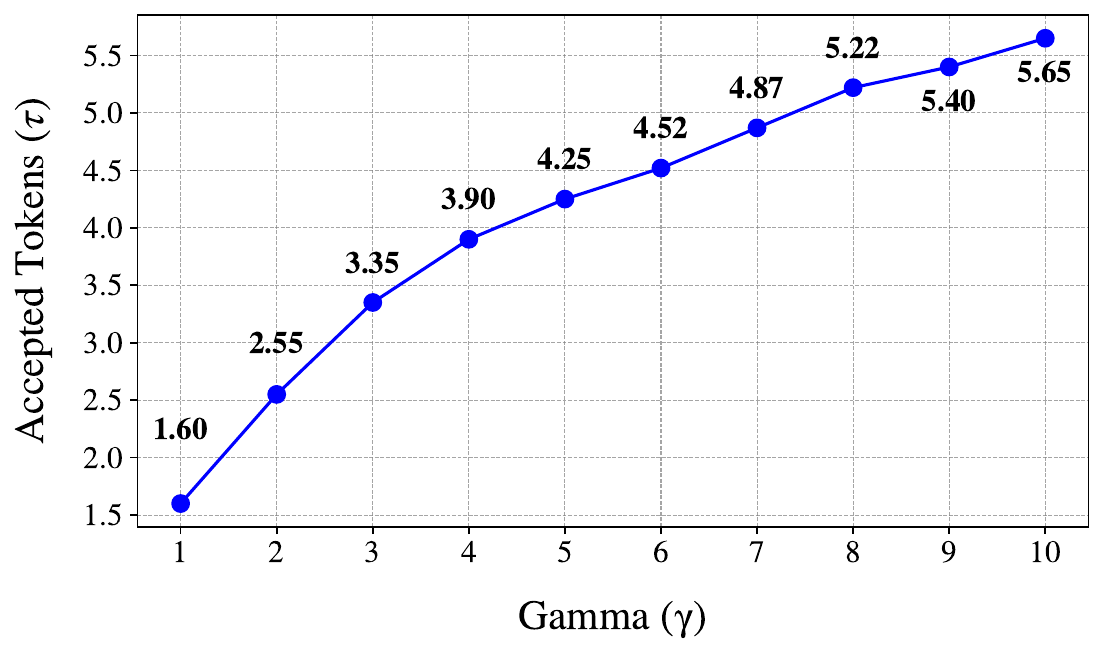}}
\caption{Average accepted tokens per copy attempt against $\gamma$ using LLaMa-8B ($|S_{\text{copyspec}}| = 10$), showcasing the correlation between $\gamma$ and the accepted tokens.} 
\label{fig:acceptance_gamma_llama8B_he}
\end{center}
% \vspace{-2cm}
\end{figure}

Furthermore, we examine the effect of $\gamma$ on $\tau$ (the average number of tokens accepted). Figure~\ref{fig:acceptance_gamma_llama8B_he} illustrates the average number of tokens accepted per attempt on HumanEval using the LLaMA3.1-8B model. We observe an interesting pattern: as $\gamma$ increases, the average number of tokens accepted per copying attempt also increases, indicating that each attempt becomes more precise. However, this comes at the cost of fewer overall copying attempts, as demonstrated in Figure~\ref{fig:speedup_gamma_llama8B_humaneval}.

This finding is particularly relevant for integrating our technique into various speculative decoding frameworks. If a framework already accepts a high number of tokens per attempt, our technique remains advantageous by increasing $\gamma$, enabling more tokens to be copied with each attempt.

\section{Analyses}
\label{sec:analyses}

\begin{table*}[ht]
\centering
\resizebox{\textwidth}{!}{
\begin{tabular}{l
                c c c c
                c c c c
                c c c c
               }
\toprule
\textbf{Variant} 
& \multicolumn{4}{c}{\textbf{Turn 1}} 
& \multicolumn{4}{c}{\textbf{Turn 2}}
& \multicolumn{4}{c}{\textbf{Turn 3}} \\
\cmidrule(lr){2-5} \cmidrule(lr){6-9} \cmidrule(lr){10-13}
& \textbf{Copied} & \textbf{Tokens/Sec} & \textbf{$\bm{\tau_1}$} & \textbf{$\bm{\tau_2}$}
& \textbf{Copied} & \textbf{Tokens/Sec} & \textbf{$\bm{\tau_1}$} & \textbf{$\bm{\tau_2}$}
& \textbf{Copied} & \textbf{Tokens/Sec} & \textbf{$\bm{\tau_1}$} & \textbf{$\bm{\tau_2}$} \\
\midrule

Base Model   & --      & 10.25$\pm0.01$         & -- & -- 
            & --      & 10.17$\pm0.01$         & -- & -- 
            & --      & 8.68$\pm0.01$         & -- & -- \\

CopySpec ($\gamma=3$)   & 5.76\%      & 10.13$\pm0.01$         & 0.58 & -- 
            & 44.17\%      & 15.72$\pm0.01$         & 4.90 & --
            & 82.79\%      & 21.89$\pm0.01$         & 7.67 & -- \\

CopySpec ($\gamma=5$)   & 1.01\%      & 9.91$\pm0.02$         & 0.72 & -- 
            & 40.67\%      & 14.79$\pm0.01$         & 6.96 & --
            & 82.78\%      & 21.39$\pm0.02$         & 8.70 & -- \\

Spec. Dec.
            & --       & 13.47$\pm0.02$         & -- & 2.55    
            & --       & 12.99$\pm0.03$         & -- & 2.31    
            & --       & 11.27$\pm0.01$         & -- & 2.75 \\

Spec. Dec. + Copy ($\gamma=3$)
            & 2.59\%       & 13.09$\pm0.02$         & 0.60 & 2.52    
            & 41.70\%       & 16.37$\pm0.04$         & 5.85 & 1.86    
            & 81.81\%       & 21.23$\pm0.04$         & 7.70 & 2.39 \\

Spec. Dec. + Copy ($\gamma=5$)
            & 0.49\%       & \textbf{13.67}$\pm0.03$         & 0.90 & 2.55    
            & 39.26\%       & \textbf{16.59}$\pm0.03$         & 7.89 & 1.92    
            & 82.58\%       & \textbf{21.91}$\pm0.02$         & 8.71 & 2.35 \\

\bottomrule
\end{tabular}
}
\caption{Performance comparison for self-correcting tasks with the draft model generating 3 tokens at a time. Qwen2.5-32B-Instruct is the target model, and Qwen2.5-7B-Instruct is the draft model. The Base Model averages 9.76 TPS, while Spec. Dec. + CopySpec ($\gamma=5$) averages 16.75 TPS across all three rounds. \(\tau_1\) is the average tokens accepted by CopySpec, and \(\tau_2\) is the average tokens accepted by the draft model. Self-correction leads to an improvement in accuracy from 92\% to 93\%, for more details see \cref{tab:selfcorrection} in Appendix.}
\label{tab:specdec_selfcorrection_3tokens}
\end{table*}

\subsection{Orthogonality with Vanilla Speculative Decoding}

We followed the steps outlined in \cref{sec:mergingspecdec} to integrate our technique into a vanilla speculative decoding framework, as described in \cite{leviathan2023speculative}. Based on our observations from \cref{sec:effectgamma}, we experimented with two different values of $\gamma$ (3 and 5) to analyze their impact on performance when used alongside speculative decoding. 

We integrated CopySpec into a vanilla speculative decoding framework, following the steps in \cref{sec:mergingspecdec} and the approach described in \cite{leviathan2023speculative}. Experiments with $\gamma$ values of 3 and 5, summarized in \cref{tab:specdec_results}, show significant efficiency improvements in the second turn of MT-Redundant. A $\gamma$ value of 5 achieves higher speedups in the first turn, while $\gamma = 3$ provides better TPS in the second turn, highlighting the need for task-specific tuning.

We also evaluated CopySpec with speculative decoding using drafts of 5 tokens instead of 3, with similar experiments conducted on MT-Redundant (\cref{tab:mtredundant_specdec_draft_5}, in Appendix) and with 3 and 5 draft tokens on MT-Bench (\cref{tab:mtbench_specdec_draft_3} and \cref{tab:mtbench_specdec_draft_5}, in Appendix). These results confirm that $\gamma = 5$ often outperforms $\gamma = 3$ when combined with Spec. Dec., emphasizing the importance of tuning $\gamma$ for optimal performance. The results also show that adding CopySpec to Spec. Dec. almost never leads to a decrease in performance.

When paired with \textsc{EAGLE}, our verifier lifts Vicuna-v1.3-13B
throughput on GSM8K from 62 TPS (plain EAGLE) to 142 TPS
(turn 3)—\textit{3.5$\times$} the greedy baseline
(Appendix~\ref{sec:comp_eagle}).

\subsection{Complementarity with External Drafting Frameworks}
\label{sec:disc_synergy}

Coupling a strong drafter with CopySpec delivers the fastest
runs across the board—up to \textbf{2.4$\times$} throughput on
Vicuna-v1.3-7B and \textbf{3.1$\times$} on Vicuna-v1.3-13B relative to the greedy
decoder (Table~\ref{tab:eagle_summary}).  Enlarging the speculation
window from 10 to 50 tokens still yields an additional 15–40 \% gain on
overlap-heavy tasks.

\begin{table}[h]
\centering
\footnotesize
\resizebox{0.9\columnwidth}{!}{%
\begin{tabular}{lcc}
\toprule
\textbf{Method} & \textbf{Tokens/Sec} & \textbf{Speed-up} \\
\midrule
Base model                         & $39.30 \pm 0.18$ & $1.00\times$ \\
PLD (window$=5$)                       & $51.08 \pm 0.11$ & $1.30\times$ \\
CopySpec ($\gamma\!=\!5$)              & $56.13 \pm 0.01$ & $1.43\times$ \\
EAGLE                                  & $82.53 \pm 0.18$ & $2.10\times$ \\
EAGLE + SAM-D                           & $84.75 \pm 0.05$ & $2.16\times$ \\
EAGLE + Copy ($\gamma\!=\!5$)*  & $\mathbf{94.84 \pm 0.16}$ & $\mathbf{2.41\times}$ \\
\bottomrule
\end{tabular}}
\caption{Throughput on \textbf{MT-Redundant} (\textit{0-shot}) for Vicuna-v1.3-7B.  
Speed-ups are versus the base model.  
Complete per-task results appear in Tables~\ref{tab:eagleperf}–\ref{tab:eagle_gsm}. * indicates that the speculation window was increased to 50 tokens instead of the default 10.}
\label{tab:eagle_summary}
\end{table}

Our experiments reveal three take-aways.  
First, coupling a strong drafter with span-level copying delivers the
fastest runs across the board—up to \textbf{2.9$\times$} throughput
on Vicuna-v1.3-7B and \textbf{3.6$\times$} on Vicuna-v1.3-13B relative to the
greedy decoder(Appendix~\ref{sec:comp_eagle}). Second, enlarging the speculation window from 10 to 50 tokens adds a
further 15–40 \% throughput when prompts contain large verbatim
spans (\textit{e.g.}, GSM8K turn 3 reaches 142 TPS on Vicuna-v1.3-13B).
Third, competing methods such as PLD and SAM-D trail by a wide
margin, underscoring that our lighter verifier integrates
more cleanly with external drafting.

\subsection{Effect on Reasoning}

An important aspect of our analysis is evaluating the impact of our technique on the efficiency of self-correction. To this end, we implemented a self-refine framework, where the model generates Python code and iteratively refines it in two steps, following a process similar to \cite{madaan2023selfrefineiterativerefinementselffeedback}. Details of the prompts and example outputs used in our experiments are provided in Appendix~\ref{sec:self-correcting}. \Cref{tab:specdec_selfcorrection_3tokens} presents the results of self-correction with copying and Speculative Decoding (SD).

Our technique becomes more effective in later turns as the model iterates over its prior reasoning. This is reflected in a significant rise in the percentage of copied tokens, tokens per second (TPS), and $\tau_1$, the average number of tokens accepted. Each copying attempt also becomes more precise as the model refines its reasoning and the context grows.

When combined with SD using $\gamma = 5$, our approach achieves better results across all three turns, as shown in the table. The first turn benefits most from SD due to minimal copying, while later turns gain greater advantages from copying. This highlights the complementary nature of the two techniques and their combined effectiveness in improving efficiency and performance. Notably, while the TPS of the base model decreases by 0.85\(\times\)  as context size grows, our technique reverses this trend, increasing the TPS in the last turn by 2.52\(\times\), showcasing its ability to leverage larger contexts.

We also extended our analysis to cases where the draft model generates 5 tokens at a time, as shown in \cref{tab:specdec_selfcorrection_5tokens} in the appendix. Additionally, \cref{tab:selfcorrection} confirms that the tested models improve their final accuracy, validating the effectiveness of our self-correction implementation. Note that accuracy is not reported for the second round, as it focuses solely on critiquing the model’s prior implementation. Across the entire self-correction process, we achieve TPS improvements of 63\%, 52\%, and 54\% for the Qwen2.5-7B, Qwen2.5-32B, and Qwen2.5-72B instruct models, respectively.

\section{Conclusion}

We introduced \textit{CopySpec}, a method that identifies repeated token sequences in a growing context and copies them efficiently without additional GPU memory or significant cost. Using a rolling hash for \(\gamma\) tokens, \textit{CopySpec} speculates on larger token blocks to reduce redundant computation.

Results across five LLMs and datasets, including MT-Redundant, show up to a 3.08\(\times\) speed-up in second-turn inference and a 49\% boost when combined with speculative decoding, without altering output quality. Future work includes dynamically tuning \(\gamma\), refining match selection, and integrating \textit{CopySpec} with parallel decoding frameworks.

\section*{Limitations}
\label{limitations}

CopySpec provides a lightweight and effective approach for accelerating LLM inference by leveraging repeated patterns in the context. It integrates seamlessly with speculative decoding and shows consistent gains across models and tasks, particularly in multi-turn settings with incremental refinements. While MT-Redundant is designed to cover a wide range of realistic scenarios—including code revision, reasoning, summarization, and stylistic rewriting—it still assumes that the repeated content appears in close proximity and with high lexical overlap. Future work will examine how CopySpec performs under looser or cross-sentence redundancy patterns, such as those found in long-form document editing or open-ended dialogue. 

The current approach uses fixed hyperparameters for the copying window $\gamma$ and speculative block size $|S_{\text{copyspec}}|$, which may not be optimal for all settings. Although our analysis in Appendix~\ref{sec:cs} shows that $\gamma$ can be reliably selected using a small number of samples based on cosine similarity trends, more adaptive strategies that adjust $\gamma$ and block length dynamically based on local context or model uncertainty could further improve generality and efficiency. 

Finally, when multiple candidate matches exist for the last $\gamma$ tokens, the current system selects the first match encountered without regard to semantic alignment. To improve copy quality and avoid suboptimal completions, future work will explore context-aware ranking mechanisms that select the most useful copy targets based on semantic or structural compatibility with the ongoing generation.
\section*{Ethical Considerations}

Our proposed method, \textit{CopySpec}, is intended as a complementary addition to existing speculative decoding techniques and does not introduce any novel text generation of its own. Instead, it operates by selecting and copying text from context, with the final output verified by the main LLM used in the respective pipeline. Consequently, any potential biases or hallucinations in the output are attributable to the choice of the main LLM, which remains an independent and external component. While CopySpec inherently reduces hallucination risk by copying existing context, it may still propagate hallucinated content if such content exists in the context itself. The responsibility for verifying such content again lies with the main LLM employed in the speculative decoding process.

\bibliography{custom}

\appendix

\section{Gamma ($\gamma$) and Semantic Implications}
\label{sec:cs}
In our framework, the generation speed of CopySpec is intricately tied to the choice of $\gamma$, which governs the length of the left context used to identify repeated sequences. The selection of an optimal $\gamma$ is critical, as it directly impacts the model’s ability to efficiently reuse tokens from the context, thereby accelerating generation. A carefully chosen $\gamma$ strikes a balance between providing sufficient contextual information for accurate copying and avoiding unnecessary computational overhead. 

If $\gamma$ is too small (e.g., $\gamma$ = 1), the context provides insufficient information to reliably identify repetitions, resulting in missed reuse opportunities and slower generation.  Conversely, when $\gamma$ is too large, the excessive context introduces redundancy and dilutes the immediate semantic relevance. While the acceptance rate may increase, the total number of tokens generated per second decreases because the model spends more time processing generate tokens itself and fewer tokens are copied in practice.

The challenge, therefore, lies in finding an optimal $\gamma$ that maximizes copying attempts while minimizing computational overhead. A well-chosen $\gamma$ ensures that the context is both semantically focused and computationally efficient, enabling the Copy mechanism to fully exploit repeated patterns in the generation process. This tradeoff underscores the importance of systematically tuning $\gamma$ to achieve the best performance across datasets.

%%%%%%%%%%%%%%%%%%%%%%%%%%%%%%%%%%%%%%%%%%%%%%%%%%%%%%%%%%%%%%%%%%%%%%%%%%%%%%%
%%%%%%%%%%%%%%%%%%%%%%%%%%%%%%%%%%%%%%%%%%%%%%%%%%%%%%%%%%%%%%%%%%%%%%%%%%%%%%%

% \subsection{Evaluation of Context Patterns}

To measure the semantic alignment between a token \( w \) and its left-\(\gamma\) token context, we fine-tuned the token embeddings using a \textbf{left-\(\gamma\) skip-gram} model, a modification of the traditional skip-gram approach. Unlike the standard skip-gram model, which maximizes the probability of a target word given a symmetric context window, our approach considers only the preceding \(\gamma\) tokens as context.  

Formally, instead of maximizing the probability \( \prod_{(w, C) \in D} P(w | C) \), where \( C \) represents a symmetric context window around the word \( w \), our \textbf{left-\(\gamma\) skip-gram} model is trained to maximize \( \prod_{(t, C_{\text{left } \gamma}) \in D} P(t | C_{\text{left } \gamma}) \), where \( C_{\text{left } \gamma} \) consists only of the last \(\gamma\) tokens in the sequence to predict the next token \( t \). This ensures that the learned embeddings capture dependencies in a unidirectional manner, aligning with the way generative models process text.  

By structuring the model in this way, we aim to quantify how much semantic meaning from the left-\(\gamma\) tokens contributes to predicting the next token. Cosine Similarity is particularly well-suited for evaluating the semantic alignment between the left-\(\gamma\) token context and the next token because it captures the directional similarity between their vector representations, regardless of magnitude. Since word embeddings encode semantic meaning in a high-dimensional space, CS provides a robust way to measure how well the left context conveys predictive information about the next token. Unlike Euclidean Distance, CS ensures that we focus solely on semantic coherence rather than raw frequency effects. This is crucial for CopySpec, as effective token reuse depends on the ability to recognize when a sequence of past tokens is not just lexically repeated but also semantically relevant to the next token. By analyzing trends in CS across different \(\gamma\)-values, we can assess whether increasing the context length improves meaningful copying or merely introduces redundant information, thereby helping us fine-tune \(\gamma\) for optimal efficiency.

The cosine similarity (CS) is computed as:
\[
\text{CS}(\vec{v}_{C_{\text{left } \gamma}}, \vec{v}_t) = \frac{\vec{v}_{C_{\text{left } \gamma}} \cdot \vec{v}_t}{\|\vec{v}_{C_{\text{left } \gamma}}\| \|\vec{v}_t\|}.
\]

Here, \(\vec{v}_{C_{\text{left } \gamma}} = \frac{1}{\gamma} \sum_{i=1}^\gamma \vec{v}_{t_i}\) represents the average embedding of the most recent \(\gamma\) tokens, where \(\{t_i\}_{i=1}^\gamma\) are the embeddings of the last \(\gamma\) tokens in the context.

\begin{figure*}[h!]
\centering
\begin{minipage}{0.45\textwidth}
    \centering
    \includegraphics[width=\textwidth]{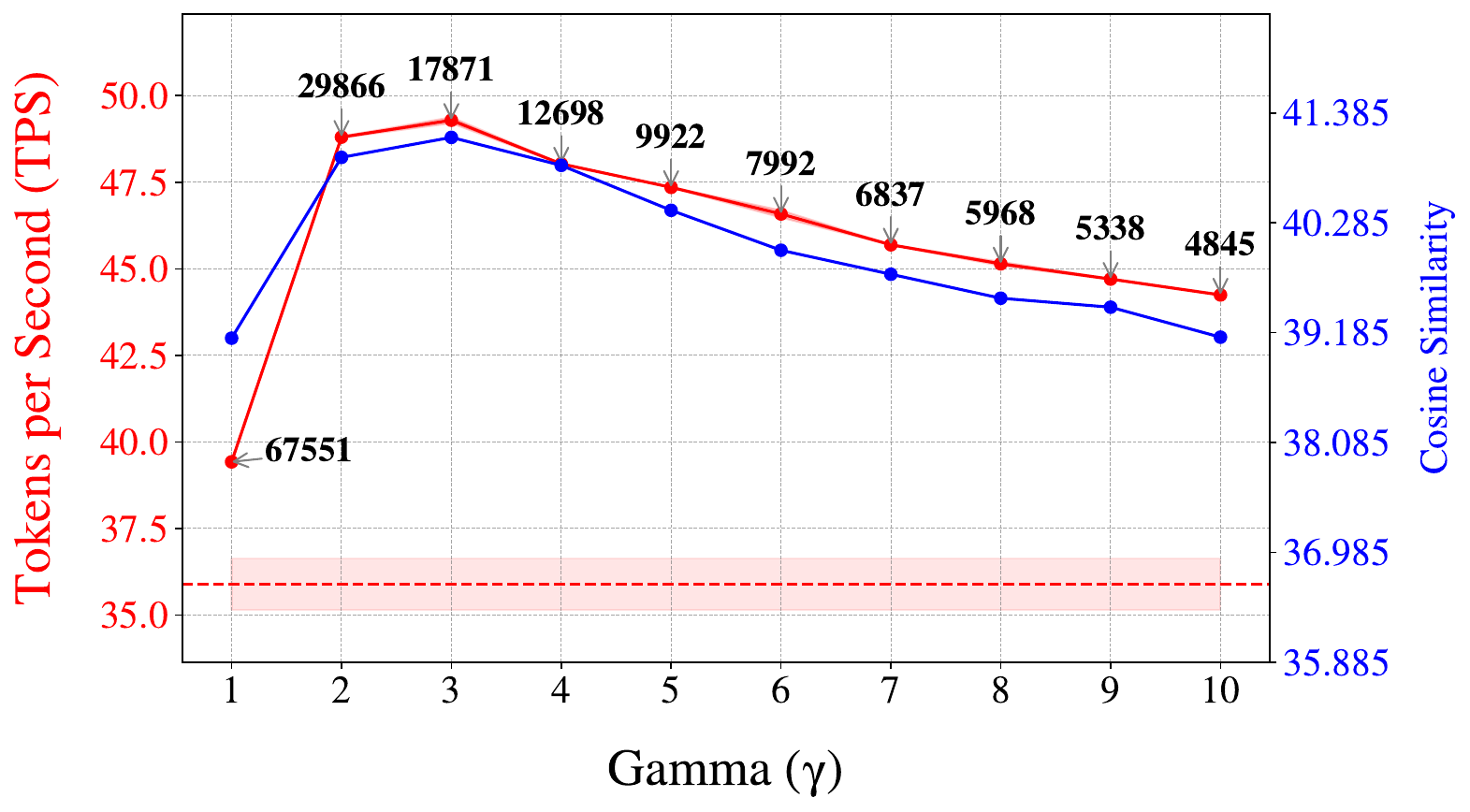} % Replace with your MT-Bench plot
\end{minipage} \hfill
\begin{minipage}{0.45\textwidth}
    \centering
    \includegraphics[width=\textwidth]{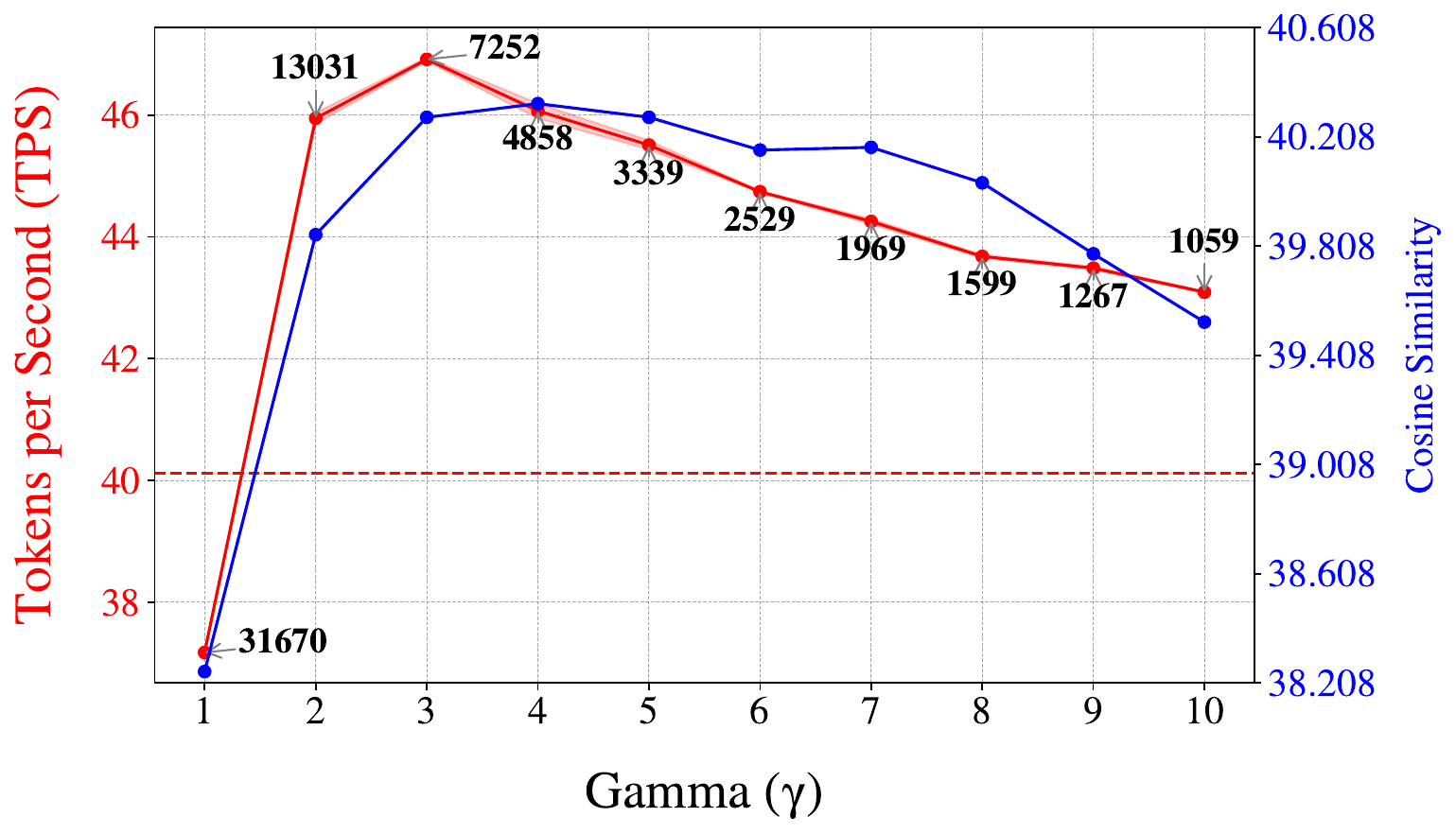} % Replace with your MT-Redundant plot
\end{minipage}
\caption{We use Qwen2.5-7B on both MT-Bench and MT-Redundant dataset. Cosine Similarity and Tokens per Second trends as a function of $\gamma$. The blue line indicates the Cosine Similarity, showing semantic alignment across varying $\gamma$-token contexts. The red line illustrates the Tokens per Second, reflecting generation speed. $\gamma$ denotes the number of tokens considered in the context for each measurement. The left plot shows MT-Bench, and the right plot shows MT-Redundant.}
\label{fig:cs-bench-redundant}
\end{figure*}

To validate our intuitions, we conducted experiments to analyze the relationship between \(\gamma\) (context length) and semantic alignment. Figure~\ref{fig:cs-bench-redundant} illustrates the trends in Cosine Similarity and generation speed (TPS) as \(\gamma\) varies.

By measuring Cosine Similarity and generation speed across varying $\gamma$-token contexts, we provide empirical evidence that fine-tuning \textbf{left-\(\gamma\) skip-gram} model for the best $\gamma$ is essential for maximizing efficiency. Future work can explore adaptive strategies that dynamically adjust $\gamma$ in the same hashmap based on context complexity, further optimizing the balance between copying effectiveness and computational cost.

%%%%%%%%%%%%%%%%%%%%%%%%%%%%%%%%%%%%%%%%%%%%%%%%%%%%%%%%%%%%%%%%%%%%%%%%%%%%%%%
%%%%%%%%%%%%%%%%%%%%%%%%%%%%%%%%%%%%%%%%%%%%%%%%%%%%%%%%%%%%%%%%%%%%%%%%%%%%%%%

%\section{Appendix: Copying and Speculative Decoding with Truncated KV States}

\section{Copying and Speculative Decoding with Truncated KV States}
\label{sec:kv_caching_appendix}

This appendix describes how our framework integrates a \emph{copying mechanism} with \emph{speculative decoding}, including details on partial acceptance, key-value (KV) cache truncation.

\subsection{Notation and Variables}

\paragraph{Sequence $X_{1:t}$.}
Let $X_{1:t}$ be the currently accepted sequence of $t$ tokens. Generating a new token moves us to position $t+1$.

\paragraph{Dictionary $\mathcal{D}$.}
$\mathcal{D}$ records repeated $\gamma$-length substrings and their earlier occurrences. If $X_{t-\gamma+1 : t}$ appears in $\mathcal{D}$, we may copy subsequent tokens from that match.

\paragraph{Subsequence length $\gamma$.}
We use $\gamma$ tokens to detect repeats. That is, the last $\gamma$ tokens, $s = X_{t-\gamma+1 : t}$, determine if a copy event is possible.

\paragraph{Match location $p$.}
If $\mathcal{D}$ indicates $X_{t-\gamma+1 : t}$ appears at position $p$, we attempt to copy tokens starting from $p + \gamma$.

\paragraph{Chunk size $m$ (copying).}
When a match is found, we form a copied chunk 
\[
    \widetilde{X}_{1:m} \;=\; 
    \bigl(\widetilde{x}_{1},\dots,\widetilde{x}_{m}\bigr)
    \;=\;
    X_{\,p+\gamma :\, p+\gamma+m-1}.
\]

\paragraph{Draft limit $\delta$ (speculative).}
If copying is not used, we let the draft model propose up to $\delta$ tokens:
\[
    \widehat{X}_{1:\delta} 
    \;=\;
    \bigl(\widehat{x}_{1},\dots,\widehat{x}_{\delta}\bigr).
\]

\paragraph{Acceptance and Draft Models.}
The target model 
\(
    p_{\mathrm{target}}(\cdot\mid X_{1:n})
\)
decides whether each new token is accepted, while the draft model 
\(
    p_{\mathrm{draft}}(X_{t}\mid X_{1:n})
\)
only proposes tokens that must still pass $p_{\mathrm{target}}$’s acceptance criterion.

\paragraph{Index $i$.}
In both copying and drafting, we iterate over newly proposed tokens with an index 
\(
    i \in \{1,\dots,m\}
\)
or 
\(
    i \in \{1,\dots,\delta\}.
\)

\paragraph{Accepted count $k$.}
Out of the $m$ (copied) or $\delta$ (drafted) tokens, only $k \le m$ or $k \le \delta$ may be accepted under $p_{\mathrm{target}}$. Rejected tokens are removed, and the key-value states are truncated to retain only 
\(
    X_{1 : t + k}.
\)

\subsection{Acceptance Criterion and KV Truncation}

Any new token $x_{t+i}$ must pass an acceptance criterion under $p_{\mathrm{target}}$; for example, at \emph{temperature} $0$, we only accept it if it is the \texttt{argmax} of the target model's conditional distribution. If the token fails, we reject it (and all subsequent tokens in the same chunk) and \emph{roll back} to $X_{1 : t+i-1}$. 

Each layer $\ell$ of the target model stores key-value tensors $(\mathbf{K}_\ell,\mathbf{V}_\ell)$ up to the final accepted token. If $k < i-1$ tokens in a chunk are accepted, we truncate $(\mathbf{K}_\ell,\mathbf{V}_\ell)$ to $t + k$ positions, ensuring the model remains consistent with the final accepted sequence.

\subsection{Integrated Generation Procedure}

Below is a single pseudocode listing that combines both \emph{copying} and \emph{speculative decoding}.

\begin{enumerate}
    \item \textbf{Check for a Copy Opportunity:}
    \begin{enumerate}
        \item Let $s = X_{t-\gamma+1 : t}$ be the most recent $\gamma$ tokens of the accepted sequence $X_{1:t}$.
        \item Check if $s$ is in $\mathcal{D}$ (the dictionary of repeats). 
            \begin{itemize}
                \item If no match exists, go to Step~\ref{step:specdec}.
            \end{itemize}
        \item Otherwise, let $p$ be the first occurrence in $\mathcal{D}(s)$ satisfying $p + \gamma - 1 < t - \gamma + 1$ (ensuring no overlap).
        \item Form a candidate chunk of length $m$: 
        \[
            \widetilde{X}_{1 : m}
            \;=\;
            X_{\,p+\gamma :\, p+\gamma + m - 1}.
        \]
        \item Initialize $k = 0$, which tracks how many tokens from $\widetilde{X}_{1:m}$ are ultimately accepted.
    \end{enumerate}

    \item \textbf{Attempt to Copy:}
    \begin{enumerate}
        \item For $i = 1$ to $m$:
        \begin{itemize}
            \item Evaluate $\widetilde{x}_{i}$ (from $\widetilde{X}_{1 : m}$) with the target model:
            \[
               p_{\mathrm{target}}\bigl(\,X_{t} \mid X_{1 : t + i - 1}\bigr).
            \]
            \item If $\widetilde{x}_{i}$ passes the acceptance criterion (e.g.\ it is the \texttt{argmax} if temperature = 0), set $k \gets k + 1$; otherwise, \emph{reject} $\widetilde{x}_{i}$ and break out of this loop.
        \end{itemize}
        \item If $k < m$:
        \begin{itemize}
            \item The final sequence is now $X_{1 : t + k}$, which means only the first $k$ tokens from $\widetilde{X}_{1:m}$ (i.e.\ $\widetilde{x}_{1}, \dots, \widetilde{x}_{k}$) are accepted.
            \item Truncate the target model's KV Cache states for all layers to length $t + k$ to discard any rejected tokens beyond position $t + k$.
        \end{itemize}
        \item Otherwise, if $k = m$, then all $m$ copied tokens are fully accepted, making $X_{1 : t + m}$ the new final sequence.
        \item \textbf{Update} $\mathcal{D}$ with any newly formed $\gamma$-subsequences ending at positions $t + j$ for $1 \le j \le k$.
    \end{enumerate}

    \item \label{step:specdec} \textbf{Speculative Decoding:}
    \begin{enumerate}
        \item If no copying occurred, generate $\delta$ tokens from the draft model:
        \[
            \widehat{X}_{1 : \delta}
            \;\sim\; 
            p_{\mathrm{draft}}\!\bigl(\,X_{t} \mid X_{1 : t}\bigr).
        \]
        \item Let $k = 0$. For $i = 1$ to $\delta$:
        \begin{itemize}
            \item Evaluate $\widehat{x}_{i}$ (from $\widehat{X}_{1 : \delta}$) using
            \[
               p_{\mathrm{target}}\bigl(\,X_{t} \mid X_{1 : t + i - 1}\bigr).
            \]
            \item If accepted, increment $k$. If rejected, break immediately.
        \end{itemize}
        \item If $k < \delta$:
        \begin{itemize}
            \item Only $\widehat{x}_{1}, \dots, \widehat{x}_{k}$ are accepted, so the final sequence is $X_{1 : t + k}$.
            \item Truncate the target model's and draft model's KV Cache states to reflect $X_{1 : t + k}$ only.
        \end{itemize}
        \item If $k = \delta$, the entire draft $\widehat{X}_{1 : \delta}$ is accepted, making $X_{1 : t + \delta}$ the new final sequence.
        \item \textbf{Update} $\mathcal{D}$ with any newly formed $\gamma$-length subsequences up to position $t + k$.
    \end{enumerate}

    \item \textbf{Repeat}: Increase $t$ by the number of accepted tokens (either $k$, $m$, or $\delta$) in this iteration. Continue until a stopping criterion (e.g.\ end-of-text token) is encountered.
\end{enumerate}

\textbf{Discussion of Truncation:}  
Whenever fewer than $m$ (in copying) or $\delta$ (in drafting) tokens are accepted, we \emph{roll back} to the accepted prefix. The target model’s key-value memory is truncated accordingly to reflect $X_{1:t + k}$. Thus, any rejected tokens do not affect the final context or the KV states.

\section{Baseline Comparisons and EAGLE}
\label{sec:comp_eagle}

The acceleration techniques discussed in the related works include PLD and PLD-copy \citep{saxena2023prompt,somasundaram2024pldacceleratingllminference}, PPD \citep{chen2024hardwareawareparallelpromptdecoding}, look-ahead decoding \citep{fu2024break}, token recycling \citep{luo2024turningtrashtreasureaccelerating}, and speculative decoding (SAM) \citep{hu2024samdecodingspeculativedecoding}. Of these, only PLD, PPD, and SAM have official publicly available code. Using the reference implementation from Somasundaram \emph{et al.}, we were able to reproduce the throughput improvements reported for PLD, observing comparable end-to-end latency reductions relative to our baseline. In contrast, applying the PPD codebase as described by Chen \emph{et al.} consistently yielded token-generation speeds below the unmodified decoder—i.e., worse-than-baseline performance in our environment. For SAM, we used the authors’ implementation with the number of predicted tokens ($n_{predicts}$) set to 15 and the default hyperparameters for EAGLE integration presented in \cite{hu2024samdecodingspeculativedecoding}. The other methods—e.g. token recycling—do not provide official open-source implementations, so we were unable to attempt replication of their reported results but our technique is orthogonal to theirs.

For the EAGLE experiments, we omitted key–value caching in the EAGLE heads due to implementation challenges within our copy framework. Instead, we set EAGLE to generate batches of 20 tokens at a time; after each batch, we check whether the copy mechanism can be applied. Whenever copying is possible, we favor it over EAGLE-generated tokens and continue copying until no further matches are found; if copying is not possible, we invoke EAGLE again to generate the next 20-token batch, repeating this generate-then-copy cycle until the desired output length is reached.

For prompt-led decoding (PLD), we fixed the prompt lookup window to three or five tokens, matching the configuration used for CopySpec. All experiments were conducted with Vicuna-1.3 in its 7B and 13B variants, since the models presented in our paper lack pretrained EAGLE weights and we experienced difficulties using the EAGLE weights with LLaMa 3.1, as the model was broken and failed to generate the correct text.

We present the results in Table~\ref{tab:eagleperf} and Table~\ref{tab:eagle_gsm}. We draw the following observations:

\begin{enumerate}[leftmargin=*,nosep]
  \item \textbf{EAGLE + CopySpec synergy}: Coupling a strong drafter (EAGLE) with span-level copying consistently yields the fastest runs—up to \emph{2.9×} speed-up on 7 B and \emph{3.6×} on 13 B relative to the plain decoder.
  \item \textbf{Speculation window matters}: Expanding the draft from 10 to 50 tokens (“*”) adds a further 15–40 \% throughput, especially on tasks with high intra-prompt overlap (GSM8K: +188\% on 7 B; +264\% on 13 B).
  \item \textbf{CopySpec alone is modest}: Without a drafter, CopySpec delivers only 40–60 \% of the EAGLE-backed gains, showing that copying and multi-token drafting are complementary.
  \item \textbf{PLD and SAM-D lag}: Prompt lookup (PLD) and SAM-D trail by a wide margin—often not surpassing plain EAGLE—hinting that our technique adds a smaller overhead and thus provides better orthogonality with other speculative decoding frameworks.
  \item \textbf{Task-level variance}: Benchmarks that reward verbatim reuse (MT-Redundant, GSM8K) benefit most, while abstractive CNN/DM summaries see the smallest absolute gains, highlighting content-overlap as a driver of copying efficiency.
  \item \textbf{Late-turn surge}: With a 50-token draft, EAGLE + CopySpec($\gamma$ = 5\*) jumps to 142 TPS on turn 3—3.5× the greedy baseline—showing that long drafts pay off once history is rich in overlaps.
  \item \textbf{SAM-D’s crossover}: SAM-D starts slower than EAGLE on turn 1 but overtakes it by turn 2, consistent with the O(1) suffix-extension cost claimed in the SAM paper.
\end{enumerate}

\begin{table*}[h!]
\centering
\footnotesize
\resizebox{\textwidth}{!}{%
\begin{tabular}{l l c c c c c}
\toprule
\textbf{Model} & \textbf{Variant} & \textbf{MT-Redundant} & \textbf{CNN/DM} & \textbf{GSM8K} & \textbf{MT-Bench} & \textbf{HumanEval} \\
 &     & \textit{0-shot} & \textit{0-shot} & \textit{3-turn} & \textit{0-shot} & \textit{0-shot} \\
\midrule
         \multirow{10}{*}{\rotatebox{90}{\textbf{Vicuna-v1.3-7B}}}   & EAGLE & 82.53$\pm$0.18 & 56.87$\pm$0.14 & 80.45$\pm$0.75 & 81.78$\pm$0.40 & 87.76$\pm$0.11 \\
         & EAGLE + CopySpec($\gamma=3$)& 87.00$\pm$0.33 & 60.62$\pm$0.34 & 88.27$\pm$0.56 & 82.92$\pm$0.57 & 101.96$\pm$0.82 \\
         & EAGLE + CopySpec($\gamma=5$)& 84.80$\pm$0.39 & 61.96$\pm$0.41 & 87.37$\pm$0.07 & 83.79$\pm$1.37 & 104.77$\pm$0.53 \\
         & EAGLE + CopySpec($\gamma=5$)*& \textbf{94.84}$\pm$0.16 & \textbf{65.63}$\pm$0.18 & \textbf{108.07}$\pm$1.02 & \textbf{87.39}$\pm$1.59 & \textbf{108.39}$\pm$0.96 \\
         & EAGLE + SAM-D& 84.75$\pm$0.05 & 50.04$\pm$0.02 & 85.18$\pm$0.13 & 73.51$\pm$0.66 & 90.00$\pm$0.10 \\
         & CopySpec($\gamma=3$)& 57.19$\pm$0.25 & 47.59$\pm$0.01 & 61.58$\pm$0.12 & 52.31$\pm$0.14 & 60.69$\pm$0.15 \\
         & CopySpec($\gamma=5$)& 56.13$\pm$0.01 & 42.95$\pm$0.14 & 60.30$\pm$0.30 & 50.40$\pm$0.03 & 57.00$\pm$0.05 \\
         & CopySpec($\gamma=5$)*& 55.79$\pm$0.04 & 41.18$\pm$0.08 & 61.71$\pm$0.23 & 49.02$\pm$0.07 & 54.64$\pm$0.03 \\
         & PLD(window=3)& 50.47$\pm$0.01 & 42.73$\pm$0.06 & 53.49$\pm$0.06  & 45.10$\pm$0.10 & 50.77$\pm$0.05 \\
         & PLD(window=5)& 51.08$\pm$0.11 & 44.45$\pm$0.04 & 56.93$\pm$0.05 & 45.05$\pm$0.12 & 51.50$\pm$0.09 \\
            & Base model & 39.30$\pm$0.18 & 31.36$\pm$0.01 & 37.48$\pm$0.11 & 39.48$\pm$0.09 & 40.40$\pm$0.09 \\
\midrule
         \multirow{10}{*}{\rotatebox{90}{\textbf{Vicuna-v1.3-13B}}}   & EAGLE & 62.31$\pm$0.18 & 43.23$\pm$0.25 & 68.35$\pm$0.29 & 62.83$\pm$0.42 & 69.34$\pm$0.45 \\
         & EAGLE + CopySpec($\gamma=3$)& 65.44$\pm$0.63 & 45.81$\pm$0.30 & 70.39$\pm$0.25 & 64.33$\pm$0.56 & 69.97$\pm$0.22 \\
         & EAGLE + CopySpec($\gamma=5$)& 66.11$\pm$0.74 & 45.08$\pm$0.52 & 71.88$\pm$0.48 & 63.58$\pm$0.99 & 73.18$\pm$0.97 \\
         & EAGLE + CopySpec($\gamma=5$)*& \textbf{73.20}$\pm$0.34 & \textbf{46.94}$\pm$0.21 & \textbf{82.27}$\pm$0.75 & \textbf{65.52}$\pm$0.78 & \textbf{74.16}$\pm$0.68 \\
         & EAGLE + SAM-D& 50.85$\pm$0.13 & 27.63$\pm$0.02 & 47.31$\pm$0.01 & 45.10$\pm$0.01 & 50.17$\pm$0.03 \\
         & CopySpec($\gamma=3$)& 34.12$\pm$0.05 & 25.18$\pm$0.06 & 34.50$\pm$0.08 & 29.27$\pm$0.02 & 31.30$\pm$0.01 \\
         & CopySpec($\gamma=5$)& 32.96$\pm$0.06 & 22.98$\pm$0.01 & 33.49$\pm$0.05 & 28.19$\pm$0.06 & 30.67$\pm$0.02 \\
         & CopySpec($\gamma=5$)*& 32.24$\pm$0.03 & 21.42$\pm$0.01 & 33.31$\pm$0.02 & 26.96$\pm$0.05 & 29.16$\pm$0.05 \\
         & PLD(window=3)& 31.57$\pm$0.09 & 25.71$\pm$0.02 & 31.99$\pm$0.06  & 28.61$\pm$0.01 & 31.51$\pm$0.05 \\
         & PLD(window=5)& 31.10$\pm$0.02 & 25.56$\pm$0.02 & 31.77$\pm$0.02 & 27.57$\pm$0.04 & 30.10$\pm$0.05 \\
            & Base model & 23.30$\pm$0.01 & 18.42$\pm$0.04 & 22.61$\pm$0.01 & 23.48$\pm$0.07 & 24.25$\pm$0.05 \\
\bottomrule
\end{tabular}
}
\caption{Performance comparison of Vicuna-1.3(7B and 13B) under baseline, EAGLE, CopySpec, and PLD configurations measured by tokens/sec. Reported values are tokens per second (mean±std). $\gamma$ denotes CopySpec’s prompt lookup size; * indicates that the speculation window was increased to 50 tokens instead of the default 10.}
\label{tab:eagleperf}
\end{table*}

\begin{table}[ht]
\centering
\resizebox{0.48\textwidth}{!}{%
\begin{tabular}{l c c c}
\toprule
\textbf{Variant} & \textbf{Turn 1} & \textbf{Turn 2} & \textbf{Turn 3} \\
\midrule

Base Model                     & 40.16$\pm$0.14 & 37.69$\pm$0.14 & 35.42$\pm$0.07 \\
EAGLE         & 76.48$\pm$0.68 & 70.87$\pm$0.91 & 88.56$\pm$0.71 \\
EAGLE + CopySpec($\gamma=3$)         & \textbf{80.85}$\pm$0.68 & \textbf{75.04}$\pm$0.56 & 101.60$\pm$0.40 \\
EAGLE + CopySpec($\gamma=5$)                    & 78.05$\pm$0.05 & 74.27$\pm$0.03 & 101.14$\pm$0.22 \\
EAGLE + CopySpec($\gamma=5$)*& 77.94$\pm$0.98 & 73.17$\pm$0.80 & \textbf{141.60}$\pm$0.96 \\
EAGLE + SAM-D & 60.27$\pm$0.15 & 98.18$\pm$0.07 & 118.56$\pm$0.08 \\
CopySpec($\gamma=3$) & 49.89$\pm$0.24 & 41.70$\pm$0.28 & 96.85$\pm$0.44 \\
CopySpec($\gamma=5$) & 47.66$\pm$0.04 & 40.38$\pm$0.02 & 99.83$\pm$1.67 \\
CopySpec($\gamma=5$)* & 44.89$\pm$0.21 & 39.27$\pm$0.13 & 126.41$\pm$0.24 \\
PLD(window=3) & 45.74$\pm$0.10 & 56.48$\pm$0.06 & 59.05$\pm$0.02 \\
PLD(window=5) & 45.10$\pm$0.01 & 61.32$\pm$0.02 & 67.30$\pm$0.18 \\

\bottomrule
\end{tabular}
}
\caption{We show the tokens/sec by turn on GSM8K using the Vicuna-v1.3-7B model. $\gamma$ denotes CopySpec’s prompt lookup size; * indicates that the speculation window was increased to 50 tokens instead of the default 10.}
\label{tab:eagle_gsm}
\end{table}

%\section{Speed-up by category}
\section{Extra Results on MT-Redundant}
\label{sec:redundant}

This appendix presents a detailed analysis of the performance improvements achieved by the CopySpec approach compared to baseline methods. The tables provide comprehensive results across various categories and model configurations, highlighting the computational efficiency and speed-ups observed on the MT-Redundant dataset.

\subsection{Analysis of Gamma ($\gamma$) on MT-Redundant}

\begin{figure}[H]
\begin{center}
\centerline{\includegraphics[width=\columnwidth]{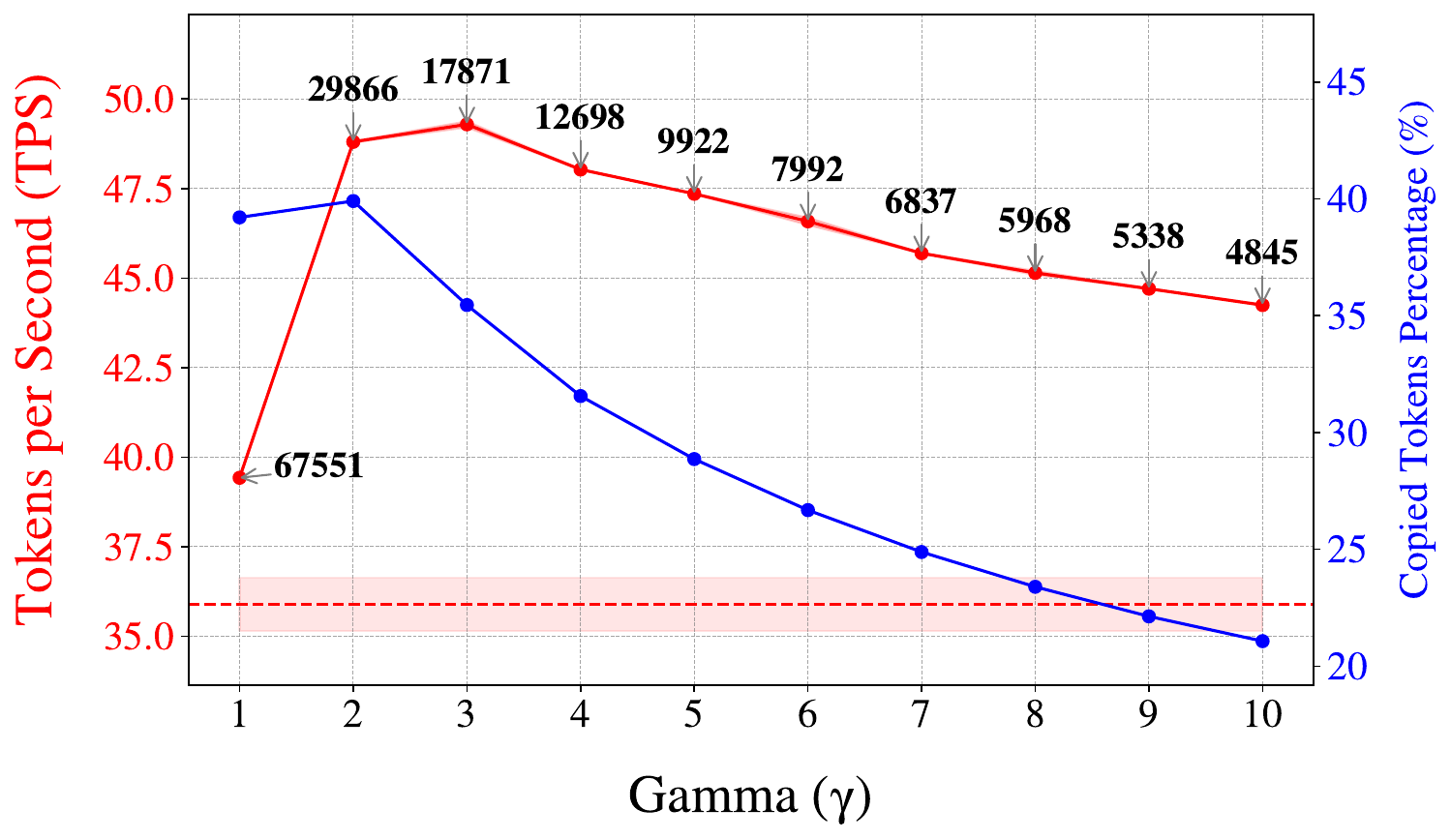}}
\caption{This figure illustrates the relationship between the copying parameter $\gamma$ and the model's performance on the MT-Redundant dataset with the LLaMa3.1-8B-Instruct model. The notations are the same as in \cref{fig:speedup_gamma_llama8B_humaneval}.}
\label{fig:speedup_gamma_llama8B_mtredundant}
\end{center}
\end{figure}

The analysis depicted in Figure~\ref{fig:speedup_gamma_llama8B_mtredundant} highlights the impact of the copying parameter $\gamma$ on both computational performance and the model's ability to reuse tokens effectively. As $\gamma$ increases, there is a notable rise in the percentage of copied tokens, demonstrating the model's improved ability to exploit repeated patterns within the context. However, this comes at the cost of reduced tokens per second (TPS) for higher $\gamma$ values, due to the increased computational overhead associated with processing larger context windows.

\subsection{Speed-up by Category on MT-Redundant}
\label{sec:extra_speedup_mtredundant}

\begin{table*}[ht]
\centering
\footnotesize
\begin{tabular}{lcccccc}
\toprule
& \multicolumn{3}{c}{\textbf{Turn 1}} & \multicolumn{3}{c}{\textbf{Turn 2}} \\
\cmidrule(lr){2-4} \cmidrule(lr){5-7}
\textbf{Category} & \textbf{Base Model} & \textbf{CopySpec} & \textbf{Speed-up} & \textbf{Base Model} & \textbf{CopySpec} & \textbf{Speed-up} \\
\midrule
Coding & 10.86 $\pm 0.01$ & \textbf{11.66} $\pm 0.02$ & 1.07 & 9.72 $\pm 0.01$ & \textbf{19.47} $\pm 0.01$ & 2.01 \\
Extraction & 10.09 $\pm 0.01$ & \textbf{13.44} $\pm 0.01$ & 1.33 & 9.80 $\pm 0.01$ & \textbf{18.17} $\pm 0.01$ & 1.85 \\
Humanities & 10.85 $\pm 0.01$ & \textbf{11.57} $\pm 0.01$ & 1.07 & 9.75 $\pm 0.01$ & \textbf{11.67} $\pm 0.01$ & 1.20 \\
Math & 11.01 $\pm 0.01$ & \textbf{12.81} $\pm 0.01$ & 1.16 & 10.05 $\pm 0.01$ & \textbf{23.18} $\pm 0.01$ & 2.31 \\
Reasoning & 10.80 $\pm 0.02$ & \textbf{12.18} $\pm 0.01$ & 1.13 & 10.05 $\pm 0.01$ & \textbf{20.17} $\pm 0.01$ & 2.01 \\
Roleplay & 10.90 $\pm 0.01$ & \textbf{11.05} $\pm 0.01$ & 1.01 & 9.93 $\pm 0.01$ & \textbf{27.80} $\pm 0.01$ & 2.80 \\
Stem & 10.90 $\pm 0.01$ & \textbf{11.50} $\pm 0.01$ & 1.06 & 9.83 $\pm 0.01$ & \textbf{14.61} $\pm 0.01$ & 1.49 \\
Writing & \textbf{10.92} $\pm 0.01$ & 10.85 $\pm 0.01$ & 0.99 & 9.94 $\pm 0.01$ & \textbf{24.51} $\pm 0.01$ & 2.46 \\
\midrule
\textbf{Average} & 10.89 $\pm 0.01$ & \textbf{11.88} $\pm 0.01$ & 1.10 & 9.88 $\pm 0.01$ & \textbf{19.52} $\pm 0.01$ & 1.98 \\
\bottomrule
\end{tabular}
\caption{Tokens per second on two turns across categories on MT-Redundant using CopySpec and Baseline with Qwen-32B-Instruct ($\gamma = 3$). Results follow the same notation as \cref{tab:category_qwen72B_redundant}.}
\label{tab:category_qwen32B_redundant_copyspec}
\end{table*}

Table~\ref{tab:category_qwen32B_redundant_copyspec} summarizes the tokens-per-second (TPS) performance for the Qwen-32B-Instruct model across two turns. The first turn reflects scenarios with minimal contextual information, while the second turn demonstrates significant gains in speed due to the larger context size and CopySpec's ability to leverage repeated token patterns effectively. Notably, categories such as Coding and Math exhibit speed-ups exceeding 2× in the second turn.

\begin{table*}[ht]
\centering
\footnotesize
\begin{tabular}{lcccccc}
\toprule
& \multicolumn{3}{c}{\textbf{Turn 1}} & \multicolumn{3}{c}{\textbf{Turn 2}} \\
\cmidrule(lr){2-4} \cmidrule(lr){5-7}
\textbf{Category} & \textbf{Base Model} & \textbf{CopySpec} & \textbf{Speed-up} & \textbf{Base Model} & \textbf{CopySpec} & \textbf{Speed-up} \\
\midrule
Coding & 43.28 $\pm 0.02$ & \textbf{47.16} $\pm 0.10$ & 1.09 & 37.48 $\pm 0.01$ & \textbf{77.39} $\pm 0.16$ & 2.06 \\
Extraction & 39.45 $\pm 0.01$ & \textbf{44.38} $\pm 0.07$ & 1.12 & 39.34 $\pm 0.01$ & \textbf{73.79} $\pm 0.15$ & 1.88 \\
Humanities & 42.94 $\pm 0.02$ & \textbf{44.73} $\pm 0.09$ & 1.04 & 36.71 $\pm 0.01$ & \textbf{46.73} $\pm 0.09$ & 1.27 \\
Math & 44.27 $\pm 0.02$ & \textbf{49.49} $\pm 0.10$ & 1.12 & 39.85 $\pm 0.01$ & \textbf{84.93} $\pm 0.43$ & 2.13 \\
Reasoning & 43.06 $\pm 0.02$ & \textbf{46.51} $\pm 0.09$ & 1.08 & 39.67 $\pm 0.03$ & \textbf{86.13} $\pm 0.14$ & 2.17 \\
Roleplay & 43.14 $\pm 0.11$ & \textbf{45.12} $\pm 0.13$ & 1.05 & 38.63 $\pm 0.02$ & \textbf{108.37} $\pm 0.18$ & 2.81 \\
Stem & 42.96 $\pm 0.04$ & \textbf{45.41} $\pm 0.07$ & 1.06 & 37.06 $\pm 0.01$ & \textbf{57.54} $\pm 0.11$ & 1.55 \\
Writing & 43.50 $\pm 0.01$ & \textbf{44.79} $\pm 0.10$ & 1.03 & 38.40 $\pm 0.01$ & \textbf{87.91} $\pm 0.12$ & 2.29 \\
\midrule
\textbf{Average} & 42.95 $\pm 0.03$ & \textbf{46.82} $\pm 0.09$ & 1.09 & 38.51 $\pm 0.01$ & \textbf{78.43} $\pm 0.17$ & 2.04 \\
\bottomrule
\end{tabular}
\caption{Tokens per second on two turns across categories on MT-Redundant using CopySpec and Baseline with Qwen-7B-Instruct ($\gamma = 3$). Results follow the same notation as \cref{tab:category_qwen72B_redundant}.}
\label{tab:category_qwen7B_redundant_copyspec}
\end{table*}

In Table~\ref{tab:category_qwen7B_redundant_copyspec}, we observe a similar trend for the Qwen-7B-Instruct model, with CopySpec consistently improving TPS across both turns. The second turn results show substantial gains in categories like Reasoning and Math, where repetitive patterns in the context are more prominent.

\begin{table*}[h]
\centering
\footnotesize
\begin{tabular}{lcccccc}
\toprule
& \multicolumn{3}{c}{\textbf{Turn 1}} & \multicolumn{3}{c}{\textbf{Turn 2}} \\
\cmidrule(lr){2-4} \cmidrule(lr){5-7}
\textbf{Category} & \textbf{Base Model} & \textbf{CopySpec} & \textbf{Speed-up} & \textbf{Base Model} & \textbf{CopySpec} & \textbf{Speed-up} \\
\midrule
Coding & 5.17 $\pm 0.01$ & \textbf{5.94} $\pm 0.01$ & 1.15 & 4.81 $\pm 0.01$ & \textbf{10.76} $\pm 0.01$ & 2.24 \\
Extraction & 4.90 $\pm 0.01$ & \textbf{5.29} $\pm 0.01$ & 1.08 & 4.80 $\pm 0.01$ & \textbf{7.60} $\pm 0.01$ & 1.58 \\
Humanities & 5.20 $\pm 0.01$ & \textbf{5.39} $\pm 0.01$ & 1.04 & 4.78 $\pm 0.01$ & \textbf{5.72} $\pm 0.01$ & 1.20 \\
Math & 5.23 $\pm 0.01$ & \textbf{5.83} $\pm 0.01$ & 1.12 & 4.89 $\pm 0.01$ & \textbf{12.58} $\pm 0.01$ & 2.57 \\
Reasoning & 5.18 $\pm 0.01$ & \textbf{5.43} $\pm 0.01$ & 1.05 & 4.92 $\pm 0.01$ & \textbf{8.49} $\pm 0.01$ & 1.73 \\
Roleplay & 5.16 $\pm 0.01$ & \textbf{5.28} $\pm 0.01$ & 1.02 & 4.93 $\pm 0.01$ & \textbf{10.01} $\pm 0.01$ & 2.03 \\
Stem & 5.21 $\pm 0.01$ & \textbf{5.43} $\pm 0.01$ & 1.04 & 4.83 $\pm 0.01$ & \textbf{6.38} $\pm 0.01$ & 1.32 \\
Writing & 5.21 $\pm 0.01$ & \textbf{5.27} $\pm 0.01$ & 1.01 & 4.82 $\pm 0.01$ & \textbf{9.48} $\pm 0.01$ & 1.97 \\
\midrule
\textbf{Average} & 
5.16 $\pm 0.01$ & 
\textbf{5.48} $\pm 0.01$ & 
1.06 & 
4.85 $\pm 0.01$ & 
\textbf{8.88} $\pm 0.01$ & 
1.83 \\
\bottomrule
\end{tabular}
\caption{Tokens per second on two turns across categories on MT-Redundant using CopySpec and Baseline with LLaMa3.1-70B-Instruct ($\gamma = 3$). Results follow the same notation as \cref{tab:category_qwen72B_redundant}.}
\label{tab:category_llama70B_redundant_copyspec}
\end{table*}

Table~\ref{tab:category_llama70B_redundant_copyspec} presents the results for the LLaMa3.1-70B-Instruct model. Here, the impact of CopySpec is evident, especially in the second turn, with speed-ups reaching over 2× in categories such as Math. These results highlight the scalability of CopySpec across models of varying sizes.

\begin{table*}[ht]
\centering
\footnotesize
\begin{tabular}{lcccccc}
\toprule
& \multicolumn{3}{c}{\textbf{Turn 1}} & \multicolumn{3}{c}{\textbf{Turn 2}} \\
\cmidrule(lr){2-4} \cmidrule(lr){5-7}
\textbf{Category} & \textbf{Base Model} & \textbf{CopySpec} & \textbf{Speed-up} & \textbf{Base Model} & \textbf{CopySpec} & \textbf{Speed-up} \\
\midrule
Coding & 36.80 $\pm 0.06$ & \textbf{44.31} $\pm 0.07$ & 1.20 & 34.61 $\pm 0.01$ & \textbf{66.14} $\pm 0.10$ & 1.91 \\
Extraction & 35.49 $\pm 0.01$ & \textbf{46.27} $\pm 0.08$ & 1.30 & 33.78 $\pm 0.01$ & \textbf{71.84} $\pm 0.07$ & 2.13 \\
Humanities & 37.31 $\pm 0.01$ & \textbf{40.66} $\pm 0.23$ & 1.09 & 33.90 $\pm 0.01$ & \textbf{40.01} $\pm 0.06$ & 1.18 \\
Math & 37.02 $\pm 0.07$ & \textbf{52.60} $\pm 0.08$ & 1.42 & 34.94 $\pm 0.05$ & \textbf{64.90} $\pm 0.07$ & 1.86 \\
Reasoning & 36.83 $\pm 0.01$ & \textbf{53.24} $\pm 0.01$ & 1.45 & 34.77 $\pm 0.04$ & \textbf{60.76} $\pm 0.09$ & 1.75 \\
Roleplay & 36.85 $\pm 0.02$ & \textbf{40.85} $\pm 0.11$ & 1.11 & 34.70 $\pm 0.02$ & \textbf{64.18} $\pm 0.13$ & 1.85 \\
Stem & 37.28 $\pm 0.04$ & \textbf{41.01} $\pm 0.10$ & 1.10 & 34.49 $\pm 0.06$ & \textbf{45.01} $\pm 0.09$ & 1.31 \\
Writing & 36.94 $\pm 0.02$ & \textbf{39.87} $\pm 0.10$ & 1.08 & 33.87 $\pm 0.02$ & \textbf{48.01} $\pm 0.09$ & 1.42 \\
\midrule
\textbf{Average} & 
36.81 $\pm 0.03$ & 
\textbf{44.85} $\pm 0.10$ & 
1.22 & 
34.38 $\pm 0.03$ & 
\textbf{57.61} $\pm 0.09$ & 
1.67 \\
\bottomrule
\end{tabular}
\caption{Tokens per second on two turns across categories on MT-Redundant using CopySpec and Baseline with LLaMa3.1-8B-Instruct ($\gamma = 3$). Results follow the same notation as \cref{tab:category_qwen72B_redundant}.}
\label{tab:category_llama8B_redundant_copyspec}
\end{table*}

The findings for the LLaMa3.1-8B-Instruct model are detailed in Table~\ref{tab:category_llama8B_redundant_copyspec}. The speed-ups in this case are slightly lower compared to larger models but still demonstrate consistent improvements across all categories, with notable efficiency gains in the second turn.

\subsection{Merging with Speculative Decoding on MT-Redundant}
\label{sd_redundant}
\begin{table*}[t]
\centering
\resizebox{\textwidth}{!}{%
\begin{tabular}{lcccccccc}
\toprule
& \multicolumn{4}{c}{\textbf{Turn 1}} & \multicolumn{4}{c}{\textbf{Turn 2}} \\
\cmidrule(lr){2-5} \cmidrule(lr){6-9}
\textbf{Category}
& \textbf{Base Model}
& \textbf{Spec.\ Dec.}
& \textbf{Spec.\ Dec.}
& \textbf{Spec.\ Dec. }
& \textbf{Base Model}
& \textbf{Spec.\ Dec.}
& \textbf{Spec.\ Dec.}
& \textbf{Spec.\ Dec.} \\
& & & \textbf{+ Copy ($\gamma=3$)} & \textbf{+ Copy ($\gamma=5$)} & & & \textbf{+ Copy ($\gamma=3$)} & \textbf{+ Copy ($\gamma=5$)} \\
\midrule
Coding      & 10.87 $\pm$ 0.01 & 16.09 $\pm$ 0.13 & 15.88 $\pm$ 0.05 & \textbf{16.09} $\pm$ 0.04 & 9.73 $\pm$ 0.01 & 15.77 $\pm$ 0.09 & 22.02 $\pm$ 0.01 & \textbf{22.50} $\pm$ 0.01 \\
Extraction  & 10.09 $\pm$ 0.01 & 14.20 $\pm$ 0.09 & 15.12 $\pm$ 0.09 & \textbf{15.26} $\pm$ 0.01 & 9.79 $\pm$ 0.01 & 15.17 $\pm$ 0.08 & 18.41 $\pm$ 0.05 & \textbf{18.45} $\pm$ 0.05 \\
Humanities  & 10.85 $\pm$ 0.01 & 12.39 $\pm$ 0.10 & \textbf{12.52} $\pm$ 0.03 & 12.50 $\pm$ 0.01 & 9.75 $\pm$ 0.01 & 12.39 $\pm$ 0.07 & 13.01 $\pm$ 0.04 & \textbf{13.05} $\pm$ 0.01 \\
Math        & 11.01 $\pm$ 0.01 & 17.61 $\pm$ 0.10 & 17.68 $\pm$ 0.06 & \textbf{18.10} $\pm$ 0.01 & 10.05 $\pm$ 0.01 & 16.70 $\pm$ 0.11 & 24.48 $\pm$ 0.07 & \textbf{24.84} $\pm$ 0.07 \\
Reasoning   & 10.80 $\pm$ 0.02 & 13.09 $\pm$ 0.10 & 13.04 $\pm$ 0.04 & \textbf{13.21} $\pm$ 0.02 & 10.05 $\pm$ 0.01 & 14.74 $\pm$ 0.06 & 20.33 $\pm$ 0.07 & \textbf{21.12} $\pm$ 0.05 \\
Roleplay    & 10.90 $\pm$ 0.01 & 11.14 $\pm$ 0.08 & \textbf{11.19} $\pm$ 0.04 & 11.17 $\pm$ 0.02 & 9.93 $\pm$ 0.01 & 16.19 $\pm$ 0.10 & 28.43 $\pm$ 0.01 & \textbf{28.44} $\pm$ 0.27 \\
Stem        & 10.90 $\pm$ 0.01 & 13.33 $\pm$ 0.11 & 13.36 $\pm$ 0.06 & \textbf{13.45} $\pm$ 0.01 & 9.83 $\pm$ 0.01 & 14.16 $\pm$ 0.08 & 16.73 $\pm$ 0.02 & \textbf{16.95} $\pm$ 0.03 \\
Writing     & 10.92 $\pm$ 0.01 & 11.30 $\pm$ 0.08 & \textbf{11.34} $\pm$ 0.03 & 11.33 $\pm$ 0.01 & 9.94 $\pm$ 0.01 & 15.59 $\pm$ 0.12 & 25.46 $\pm$ 0.01 & \textbf{25.16} $\pm$ 0.05 \\
\midrule
\textbf{Average}
& 10.79 $\pm$ 0.01 & 13.64 $\pm$ 0.10 & 13.77 $\pm$ 0.05 & \textbf{13.89} $\pm$ 0.01 & 9.88 $\pm$ 0.01 & 15.09 $\pm$ 0.09 & 21.11 $\pm$ 0.04 & \textbf{21.31} $\pm$ 0.07 \\
\bottomrule
\end{tabular}%
}
\caption{Tokens-per-second (TPS) performance on the MT-Redundant dataset, using Qwen2.5-32B-Instruct as the target model and Qwen2.5-7B-Instruct as the draft model, where the draft model generates 5 tokens per attempt. Results are presented using the same notation as Table 3 and a $\gamma$ value of 3, highlighting the impact of varying the draft token count on computational efficiency.}
\label{tab:mtredundant_specdec_draft_5}
\end{table*}

Finally, Table~\ref{tab:mtredundant_specdec_draft_5} explores the integration of CopySpec with speculative decoding for the Qwen2.5-32B-Instruct model and Qwen2.5-7B-Instruct as the draft model. The results highlight how combining these approaches can yield even greater computational efficiency. The analysis includes varying $\gamma$ values and draft token counts, showing that optimal parameter tuning further enhances performance, particularly in multi-turn scenarios.

\section{Extra Results on MT-Bench}
\label{sec:bench}
This appendix presents a comprehensive evaluation of the CopySpec approach on the MT-Bench dataset across various configurations and categories. The results highlight the consistent improvements in tokens-per-second (TPS) performance achieved by CopySpec compared to baseline models, demonstrating its efficiency and scalability.

\subsection{Analysis of Gamma ($\gamma$) on MT-Bench}

\begin{figure*}[h]
\begin{center}
\centerline{\includegraphics[width=1.2\columnwidth]{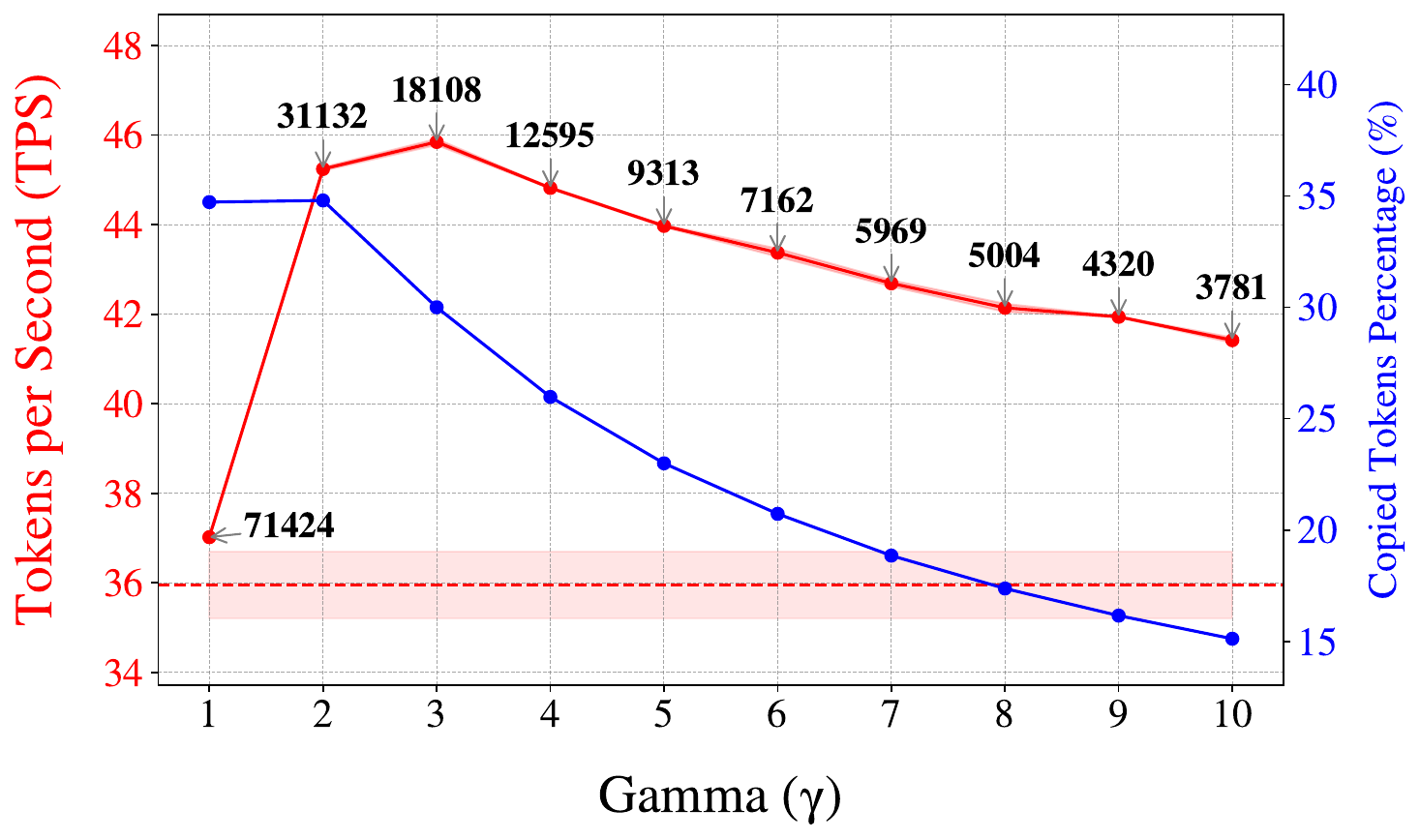}}
\caption{This figure illustrates the relationship between the copying parameter $\gamma$ and the model's performance on the MT-Bench dataset with the LLaMa3.1-8B-Instruct model. The notations are the same as in \cref{fig:speedup_gamma_llama8B_humaneval}.}
\label{fig:speedup_gamma_llama8B_mtbench}
\end{center}
\end{figure*}

\Cref{fig:speedup_gamma_llama8B_mtbench} presents a comprehensive visualization of how the copying parameter $\gamma$ affects the performance of the LLaMa3.1-8B-Instruct model on the MT-Redundant dataset. The figure captures the interplay between the percentage of tokens successfully copied, the number of copying attempts, and the resulting tokens per second (TPS).

\subsection{Speed-up by Category on MT-Bench}
\label{sec:extra_speedup_mtbench}

\begin{table*}[ht]
\centering
\footnotesize
\begin{tabular}{lcccccc}
\toprule
& \multicolumn{3}{c}{\textbf{Turn 1}} & \multicolumn{3}{c}{\textbf{Turn 2}} \\
\cmidrule(lr){2-4} \cmidrule(lr){5-7}
\textbf{Category} & \textbf{Baseline} & \textbf{CopySpec} & \textbf{Speed-up} & \textbf{Baseline} & \textbf{CopySpec} & \textbf{Speed-up} \\
\midrule
Coding & 5.12 $\pm 0.01$ & \textbf{5.62} $\pm 0.01$ & 1.10 & 4.62 $\pm 0.01$ & \textbf{7.10} $\pm 0.01$ & 1.54 \\
Extraction & 4.76 $\pm 0.01$ & \textbf{5.64} $\pm 0.01$ & 1.19 & 4.48 $\pm 0.01$ & \textbf{6.84} $\pm 0.01$ & 1.53 \\
Humanities & 5.09 $\pm 0.01$ & \textbf{5.32} $\pm 0.01$ & 1.04 & 4.54 $\pm 0.01$ & \textbf{4.98} $\pm 0.01$ & 1.10 \\
Math & 5.17 $\pm 0.01$ & \textbf{5.84} $\pm 0.01$ & 1.13 & 4.81 $\pm 0.01$ & \textbf{6.72} $\pm 0.01$ & 1.40 \\
Reasoning & 5.08 $\pm 0.01$ & \textbf{5.69} $\pm 0.01$ & 1.12 & 4.80 $\pm 0.01$ & \textbf{5.96} $\pm 0.01$ & 1.24 \\
Roleplay & 5.06 $\pm 0.01$ & \textbf{5.14} $\pm 0.01$ & 1.02 & 4.59 $\pm 0.01$ & \textbf{4.68} $\pm 0.01$ & 1.02 \\
Stem & 5.12 $\pm 0.01$ & \textbf{5.38} $\pm 0.01$ & 1.05 & 4.62 $\pm 0.01$ & \textbf{5.32} $\pm 0.01$ & 1.15 \\
Writing & 5.12 $\pm 0.01$ & \textbf{5.12} $\pm 0.01$ & 1.01 & 4.69 $\pm 0.01$ & \textbf{6.09} $\pm 0.01$ & 1.30 \\
\midrule
\textbf{Average} & 
5.07 $\pm 0.01$ & 
\textbf{5.47} $\pm 0.01$ & 
1.08 & 
4.64 $\pm 0.01$ & 
\textbf{5.96} $\pm 0.01$ & 
1.28 \\
\bottomrule
\end{tabular}
\caption{Tokens per second on two turns across categories on MT-Bench using CopySpec and Baseline with Qwen2.5-72B-Chat ($\gamma = 3$). Results follow the same notation as \cref{tab:category_qwen72B_redundant}.}
\label{tab:category_qwen72B_copyspec}
\end{table*}

Table~\ref{tab:category_qwen72B_copyspec} provides the TPS performance of Qwen2.5-72B-Chat on two turns. The speed-ups are most notable in categories such as Extraction and Coding, where repetitive patterns allow CopySpec to outperform the baseline consistently. Average speed-ups for both turns reinforce the efficiency gains achieved.

\begin{table*}[ht]
\centering
\footnotesize
\begin{tabular}{lcccccc}
\toprule
& \multicolumn{3}{c}{\textbf{Turn 1}} & \multicolumn{3}{c}{\textbf{Turn 2}} \\
\cmidrule(lr){2-4} \cmidrule(lr){5-7}
\textbf{Category} & \textbf{Base Model} & \textbf{CopySpec} & \textbf{Speed-up} & \textbf{Base Model} & \textbf{CopySpec} & \textbf{Speed-up} \\
\midrule
Coding & 10.86 $\pm 0.01$ & \textbf{11.67 $\pm 0.01$} & 1.07 & 9.73 $\pm 0.01$ & \textbf{17.03 $\pm 0.01$} & 1.75 \\
Extraction & 10.09 $\pm 0.01$ & \textbf{13.39 $\pm 0.04$} & 1.33 & 9.59 $\pm 0.01$ & \textbf{15.40 $\pm 0.04$} & 1.61 \\
Humanities & 10.86 $\pm 0.01$ & \textbf{11.56 $\pm 0.01$} & 1.06 & 9.73 $\pm 0.01$ & \textbf{11.14 $\pm 0.01$} & 1.14 \\
Math & 11.01 $\pm 0.01$ & \textbf{12.77 $\pm 0.07$} & 1.16 & 10.15 $\pm 0.01$ & \textbf{13.35 $\pm 0.03$} & 1.32 \\
Reasoning & 10.82 $\pm 0.01$ & \textbf{12.18 $\pm 0.01$} & 1.13 & 10.22 $\pm 0.01$ & \textbf{11.54 $\pm 0.01$} & 1.13 \\
Roleplay & 10.90 $\pm 0.01$ & \textbf{11.04 $\pm 0.01$} & 1.01 & 10.16 $\pm 0.01$ & \textbf{10.37 $\pm 0.01$} & 1.02 \\
Stem & 10.89 $\pm 0.01$ & \textbf{11.51 $\pm 0.01$} & 1.06 & 9.84 $\pm 0.01$ & \textbf{11.50 $\pm 0.01$} & 1.17 \\
Writing & \textbf{10.90 $\pm 0.01$} & 10.82 $\pm 0.02$ & 0.99 & 9.99 $\pm 0.01$ & \textbf{13.25 $\pm 0.01$} & 1.33 \\
\midrule
\textbf{Average} & 10.91 $\pm 0.01$ & \textbf{11.86 $\pm 0.01$} & 1.09 & 9.92 $\pm 0.01$ & \textbf{12.57 $\pm 0.01$} & 1.27 \\
\bottomrule
\end{tabular}
\caption{Tokens per second on two turns across categories on MT-Bench using CopySpec and Baseline with Qwen2.5-32B-Chat ($\gamma = 3$). Results follow the same notation as \cref{tab:category_qwen72B_redundant}.}
\label{tab:category_qwen32B_copyspec}
\end{table*}

In Table~\ref{tab:category_qwen32B_copyspec}, the performance of Qwen2.5-32B-Chat is evaluated. CopySpec achieves significant speed-ups, particularly in the second turn, where contextual repetition becomes more prevalent. Categories like Math and Writing show marked improvements, underscoring CopySpec’s ability to handle computationally intensive tasks effectively.

\begin{table*}[ht]
\centering
\footnotesize
\begin{tabular}{lcccccc}
\toprule
& \multicolumn{3}{c}{\textbf{Turn 1}} & \multicolumn{3}{c}{\textbf{Turn 2}} \\
\cmidrule(lr){2-4} \cmidrule(lr){5-7}
\textbf{Category} & \textbf{Base Model} & \textbf{CopySpec} & \textbf{Speed-up} & \textbf{Base Model} & \textbf{CopySpec} & \textbf{Speed-up} \\
\midrule
Coding & 43.04 $\pm 0.25$ & \textbf{47.22} $\pm 0.02$ & 1.10 & 37.43 $\pm 0.08$ & \textbf{60.06} $\pm 0.01$ & 1.60 \\
Extraction & 39.50 $\pm 0.06$ & \textbf{44.41} $\pm 0.01$ & 1.12 & 38.94 $\pm 0.09$ & \textbf{52.85} $\pm 0.01$ & 1.36 \\
Humanities & 43.06 $\pm 0.07$ & \textbf{44.79} $\pm 0.01$ & 1.04 & 36.82 $\pm 0.08$ & \textbf{43.05} $\pm 0.01$ & 1.17 \\
Math & 44.40 $\pm 0.16$ & \textbf{49.46} $\pm 0.12$ & 1.11 & 39.39 $\pm 0.36$ & \textbf{53.45} $\pm 0.01$ & 1.36 \\
Reasoning & 43.49 $\pm 0.36$ & \textbf{46.57} $\pm 0.01$ & 1.07 & 40.96 $\pm 0.19$ & \textbf{46.76} $\pm 0.01$ & 1.14 \\
Roleplay & 43.43 $\pm 0.05$ & \textbf{45.35} $\pm 0.01$ & 1.04 & 38.72 $\pm 0.08$ & \textbf{39.89} $\pm 0.01$ & 1.03 \\
Stem & 43.30 $\pm 0.07$ & \textbf{45.47} $\pm 0.01$ & 1.05 & 37.34 $\pm 0.09$ & \textbf{43.61} $\pm 0.01$ & 1.17 \\
Writing & 43.58 $\pm 0.06$ & \textbf{44.72} $\pm 0.01$ & 1.03 & 38.80 $\pm 0.08$ & \textbf{55.90} $\pm 0.01$ & 1.44 \\
\midrule
\textbf{Average} & 42.80 $\pm 0.13$ & \textbf{46.98} $\pm 0.03$ & 1.10 & 38.25 $\pm 0.13$ & \textbf{49.57} $\pm 0.01$ & 1.30 \\
\bottomrule
\end{tabular}
\caption{Tokens per second on two turns across categories on MT-Bench using CopySpec and Baseline with Qwen2.5-7B-Chat ($\gamma = 3$). Results follow the same notation as \cref{tab:category_qwen72B_redundant}.}
\label{tab:category_qwen7B_copyspec}
\end{table*}

Table~\ref{tab:category_qwen7B_copyspec} highlights the results for Qwen2.5-7B-Chat. While the base model already performs efficiently, CopySpec further enhances TPS, with average speed-ups exceeding 1.3× in the second turn. These results confirm that CopySpec scales well across different model sizes.

\begin{table*}[ht]
\centering
\footnotesize
\begin{tabular}{lcccccc}
\toprule
& \multicolumn{3}{c}{\textbf{Turn 1}} & \multicolumn{3}{c}{\textbf{Turn 2}} \\
\cmidrule(lr){2-4} \cmidrule(lr){5-7}
\textbf{Category} & \textbf{Base Model} & \textbf{CopySpec} & \textbf{Speed-up} & \textbf{Base Model} & \textbf{CopySpec} & \textbf{Speed-up} \\
\midrule
Coding & 5.18 $\pm 0.01$ & \textbf{5.94} $\pm 0.01$ & 1.15 & 4.79 $\pm 0.01$ & \textbf{7.63} $\pm 0.01$ & 1.59 \\
Extraction & 4.91 $\pm 0.01$ & \textbf{5.28} $\pm 0.01$ & 1.08 & 4.65 $\pm 0.01$ & \textbf{7.03} $\pm 0.01$ & 1.51 \\
Humanities & 5.21 $\pm 0.01$ & \textbf{5.39} $\pm 0.01$ & 1.04 & 4.77 $\pm 0.01$ & \textbf{5.35} $\pm 0.01$ & 1.12 \\
Math & 5.23 $\pm 0.01$ & \textbf{5.83} $\pm 0.01$ & 1.12 & 4.96 $\pm 0.01$ & \textbf{6.57} $\pm 0.01$ & 1.32 \\
Reasoning & 5.16 $\pm 0.01$ & \textbf{5.43} $\pm 0.01$ & 1.05 & 4.96 $\pm 0.01$ & \textbf{5.56} $\pm 0.01$ & 1.12 \\
Roleplay & 5.17 $\pm 0.01$ & \textbf{5.28} $\pm 0.01$ & 1.02 & 4.94 $\pm 0.01$ & \textbf{5.90} $\pm 0.01$ & 1.19 \\
Stem & 5.22 $\pm 0.01$ & \textbf{5.41} $\pm 0.01$ & 1.04 & 4.85 $\pm 0.01$ & \textbf{5.54} $\pm 0.01$ & 1.14 \\
Writing & 5.21 $\pm 0.01$ & \textbf{5.27} $\pm 0.01$ & 1.01 & 4.81 $\pm 0.01$ & \textbf{6.42} $\pm 0.01$ & 1.33 \\
\midrule
\textbf{Average} & 
5.16 $\pm 0.01$ & 
\textbf{5.48} $\pm 0.01$ & 
1.06 & 
4.84 $\pm 0.01$ & 
\textbf{6.25} $\pm 0.01$ & 
1.29 \\
\bottomrule
\end{tabular}
\caption{Tokens per second on two turns across categories on MT-Bench using CopySpec and Baseline with LLaMa3.1-70B-Instruct ($\gamma = 3$). Results follow the same notation as \cref{tab:category_qwen72B_redundant}.}
\label{tab:category_llama70B_copyspec}
\end{table*}

The performance of LLaMa3.1-70B-Instruct is detailed in Table~\ref{tab:category_llama70B_copyspec}. CopySpec achieves consistent improvements across both turns, with substantial gains in computationally intensive categories such as Coding and Extraction. These results demonstrate the robustness of CopySpec when applied to larger models.

\begin{table*}[ht]
\centering
\footnotesize
\begin{tabular}{lcccccc}
\toprule
& \multicolumn{3}{c}{\textbf{Turn 1}} & \multicolumn{3}{c}{\textbf{Turn 2}} \\
\cmidrule(lr){2-4} \cmidrule(lr){5-7}
\textbf{Category} & \textbf{Base Model} & \textbf{CopySpec} & \textbf{Speed-up} & \textbf{Base Model} & \textbf{CopySpec} & \textbf{Speed-up} \\
\midrule
Coding & 36.86 $\pm 0.01$ & \textbf{44.35} $\pm 0.06$ & 1.20 & 34.42 $\pm 0.01$ & \textbf{53.22} $\pm 0.06$ & 1.55 \\
Extraction & 35.32 $\pm 0.07$ & \textbf{46.27} $\pm 0.03$ & 1.31 & 33.71 $\pm 0.01$ & \textbf{51.48} $\pm 0.06$ & 1.53 \\
Humanities & 37.20 $\pm 0.01$ & \textbf{40.88} $\pm 0.06$ & 1.10 & 33.78 $\pm 0.02$ & \textbf{40.61} $\pm 0.05$ & 1.20 \\
Math & 36.99 $\pm 0.01$ & \textbf{52.46} $\pm 0.24$ & 1.42 & 34.96 $\pm 0.01$ & \textbf{58.47} $\pm 0.07$ & 1.67 \\
Reasoning & 36.70 $\pm 0.04$ & \textbf{53.33} $\pm 0.06$ & 1.45 & 34.76 $\pm 0.01$ & \textbf{53.86} $\pm 0.06$ & 1.55 \\
Roleplay & 36.77 $\pm 0.01$ & \textbf{40.89} $\pm 0.06$ & 1.11 & 34.56 $\pm 0.01$ & \textbf{49.16} $\pm 0.06$ & 1.42 \\
Stem & 37.19 $\pm 0.01$ & \textbf{41.06} $\pm 0.06$ & 1.10 & 34.47 $\pm 0.01$ & \textbf{41.88} $\pm 0.06$ & 1.21 \\
Writing & 36.85 $\pm 0.01$ & \textbf{39.91} $\pm 0.06$ & 1.08 & 33.78 $\pm 0.01$ & \textbf{38.72} $\pm 0.06$ & 1.15 \\
\midrule
\textbf{Average} & 
36.73 $\pm 0.08$ & 
\textbf{44.89} $\pm 0.08$ & 
1.22 $\pm 0.02$ & 
34.30 $\pm 0.01$ & 
\textbf{48.42} $\pm 0.06$ & 
1.41 \\
\bottomrule
\end{tabular}
\caption{Tokens per second on two turns across categories on MT-Bench using CopySpec and Baseline with LLaMa3.1-8B-Instruct ($\gamma = 3$). Results follow the same notation as \cref{tab:category_qwen72B_redundant}.}
\label{tab:category_llama8B_copyspec}
\end{table*}

Table~\ref{tab:category_llama8B_copyspec} evaluates LLaMa3.1-8B-Instruct. While the model size is significantly smaller, CopySpec still yields notable improvements, particularly in the second turn, where repetitive token patterns amplify the efficiency of speculative copying.

\subsection{Merging with Speculative Decoding on MT-Bench}

\begin{table*}[t]
\centering
\resizebox{\textwidth}{!}{%
\begin{tabular}{lcccccccc}
\toprule
& \multicolumn{4}{c}{\textbf{Turn 1}} & \multicolumn{4}{c}{\textbf{Turn 2}} \\
\cmidrule(lr){2-5} \cmidrule(lr){6-9}
\textbf{Category}
& \textbf{Base Model}
& \textbf{Spec.\ Dec.}
& \textbf{Spec.\ Dec.}
& \textbf{Spec.\ Dec. }
& \textbf{Base Model}
& \textbf{Spec.\ Dec.}
& \textbf{Spec.\ Dec.}
& \textbf{Spec.\ Dec.} \\
& & & \textbf{+ Copy ($\gamma=3$)} & \textbf{+ Copy ($\gamma=5$)} & & & \textbf{+ Copy ($\gamma=3$)} & \textbf{+ Copy ($\gamma=5$)} \\
\midrule
Coding      & 10.86 $\pm 0.01$ & 15.97 $\pm 0.01$ & 15.91 $\pm 0.06$ & \textbf{16.16} $\pm 0.05$ & 9.73 $\pm 0.01$ & 14.81 $\pm 0.01$ & 19.94 $\pm 0.01$ & \textbf{19.97} $\pm 0.11$ \\
Extraction  & 10.09 $\pm 0.01$ & 14.22 $\pm 0.01$ & \textbf{15.39} $\pm 0.06$ & 15.36 $\pm 0.05$ & 9.59 $\pm 0.01$ & 14.55 $\pm 0.01$ & \textbf{16.71} $\pm 0.05$ & 16.26 $\pm 0.01$ \\
Humanities  & 10.86 $\pm 0.01$ & 13.66 $\pm 0.01$ & \textbf{13.89} $\pm 0.01$ & 13.87 $\pm 0.04$ & 9.73 $\pm 0.01$ & 12.30 $\pm 0.01$ & \textbf{12.93} $\pm 0.02$ & 12.85 $\pm 0.01$ \\
Math        & 11.01 $\pm 0.01$ & 17.02 $\pm 0.03$ & 17.30 $\pm 0.01$ & \textbf{17.32} $\pm 0.02$ & 10.15 $\pm 0.01$ & 15.38 $\pm 0.01$ & 16.04 $\pm 0.06$ & \textbf{16.61} $\pm 0.02$ \\
Reasoning   & 10.82 $\pm 0.01$ & 14.02 $\pm 0.01$ & \textbf{14.34} $\pm 0.02$ & 14.26 $\pm 0.01$ & 10.23 $\pm 0.01$ & 12.99 $\pm 0.01$ & 13.18 $\pm 0.02$ & \textbf{13.42} $\pm 0.05$ \\
Roleplay    & 10.90 $\pm 0.01$ & 12.86 $\pm 0.02$ & 12.88 $\pm 0.04$ & \textbf{12.94} $\pm 0.01$ & 10.16 $\pm 0.01$ & 12.11 $\pm 0.01$ & 12.18 $\pm 0.03$ & \textbf{12.24} $\pm 0.02$ \\
Stem        & 10.89 $\pm 0.01$ & 14.29 $\pm 0.06$ & 14.36 $\pm 0.03$ & \textbf{14.47} $\pm 0.02$ & 9.84 $\pm 0.01$ & 13.13 $\pm 0.01$ & 13.71 $\pm 0.01$ & \textbf{13.77} $\pm 0.05$ \\
Writing     & 10.90 $\pm 0.01$ & 12.65 $\pm 0.02$ & 12.69 $\pm 0.02$ & \textbf{12.71} $\pm 0.03$ & 9.99 $\pm 0.01$ & 11.69 $\pm 0.01$ & \textbf{13.54} $\pm 0.03$ & 13.31 $\pm 0.01$ \\
\midrule
\textbf{Average}
& 10.79 $\pm 0.01$ & 14.34 $\pm 0.02$ & 14.60 $\pm 0.03$ & \textbf{14.64} $\pm 0.03$ & 9.93 $\pm 0.01$ & 13.37 $\pm 0.01$ & 14.78 $\pm 0.03$ & \textbf{14.80} $\pm 0.04$ \\
\bottomrule
\end{tabular}%
}
\caption{Tokens-per-second (TPS) performance on the MT-Bench dataset, using Qwen2.5-32B-Instruct as the target model and Qwen2.5-7B-Instruct as the draft model, where the draft model generates 3 tokens per attempt. Results are presented using the same notation as Table 3 and a $\gamma$ value of 3, showcasing the improvements in speed and efficiency enabled by CopySpec.}
\label{tab:mtbench_specdec_draft_3}
\end{table*}

\begin{table*}[t]
\centering
\resizebox{\textwidth}{!}{%
\begin{tabular}{lcccccccc}
\toprule
& \multicolumn{4}{c}{\textbf{Turn 1}} & \multicolumn{4}{c}{\textbf{Turn 2}} \\
\cmidrule(lr){2-5} \cmidrule(lr){6-9}
\textbf{Category} 
& \textbf{Base Model} 
& \textbf{Spec.\ Dec.} 
& \textbf{Spec.\ Dec.} 
& \textbf{Spec.\ Dec. } 
& \textbf{Base Model} 
& \textbf{Spec.\ Dec.} 
& \textbf{Spec.\ Dec.} 
& \textbf{Spec.\ Dec.} \\
& & & \textbf{+ Copy ($\gamma=3$)} & \textbf{+ Copy ($\gamma=5$)} & & & \textbf{+ Copy ($\gamma=3$)} & \textbf{+ Copy ($\gamma=5$)} \\
\midrule
Coding      & 10.86 $\pm 0.01$ & \textbf{16.09} $\pm 0.09$ & 15.89 $\pm 0.02$ & 16.06 $\pm 0.05$ & 9.73 $\pm 0.01$ & 15.72 $\pm 0.06$ & 20.08 $\pm 0.03$ & \textbf{20.22 $\pm 0.13$} \\
Extraction  & 10.09 $\pm 0.01$ & 14.28 $\pm 0.06$ & 15.08 $\pm 0.01$ & \textbf{15.20 $\pm 0.02$} & 9.59 $\pm 0.01$ & 15.46 $\pm 0.06$ & 16.89 $\pm 0.01$ & \textbf{16.93 $\pm 0.01$} \\
Humanities  & 10.86 $\pm 0.01$ & 12.41 $\pm 0.07$ & \textbf{12.52 $\pm 0.01$} & 12.45 $\pm 0.04$ & 9.73 $\pm 0.01$ & 11.67 $\pm 0.04$ & \textbf{12.08 $\pm 0.01$} & 12.02 $\pm 0.02$ \\
Math        & 11.01 $\pm 0.01$ & 17.60 $\pm 0.15$ & 17.76 $\pm 0.02$ & \textbf{17.95 $\pm 0.06$} & 10.15 $\pm 0.01$ & 16.22 $\pm 0.06$ & 16.57 $\pm 0.02$ & \textbf{17.08 $\pm 0.01$} \\
Reasoning   & 10.82 $\pm 0.01$ & \textbf{13.04} $\pm 0.01$ & 12.94 $\pm 0.02$ & 12.97 $\pm 0.10$ & 10.23 $\pm 0.01$ & 11.92 $\pm 0.06$ & 12.25 $\pm 0.01$ & \textbf{12.29 $\pm 0.04$} \\
Roleplay    & 10.90 $\pm 0.01$ & 11.15 $\pm 0.04$ & \textbf{11.18 $\pm 0.01$} & 11.14 $\pm 0.03$ & 10.16 $\pm 0.01$ & 11.09 $\pm 0.05$ & \textbf{11.11 $\pm 0.01$} & 11.13 $\pm 0.03$ \\
Stem        & 10.89 $\pm 0.01$ & 13.34 $\pm 0.07$ & 13.35 $\pm 0.04$ & \textbf{13.37 $\pm 0.04$} & 9.84 $\pm 0.01$ & 12.87 $\pm 0.05$ & \textbf{13.12 $\pm 0.02$} & 13.12 $\pm 0.03$ \\
Writing     & 10.90 $\pm 0.01$ & 11.32 $\pm 0.04$ & \textbf{11.33 $\pm 0.01$} & 11.20 $\pm 0.11$ & 9.99 $\pm 0.01$ & 10.71 $\pm 0.06$ & \textbf{11.89 $\pm 0.01$} & 11.74 $\pm 0.01$ \\
\midrule
\textbf{Average} & 10.79 $\pm 0.01$ & 13.65 $\pm 0.07$ & 13.76 $\pm 0.02$ & \textbf{13.79} $\pm 0.06$ & 9.93 $\pm 0.01$ & 13.21 $\pm$ 0.06 & 14.25 $\pm$ 0.02 & \textbf{14.32} $\pm$ 0.04 \\
\bottomrule
\end{tabular}%
}
\caption{Tokens-per-second (TPS) performance on the MT-Bench dataset, using Qwen2.5-32B-Instruct as the target model and Qwen2.5-7B-Instruct as the draft model, where the draft model generates 5 tokens per attempt. Results are presented using the same notation as Table 3 and a $\gamma$ value of 3, illustrating the scalability and efficiency of CopySpec under varied settings.}
\label{tab:mtbench_specdec_draft_5}

\end{table*}

Finally, Table~\ref{tab:mtbench_specdec_draft_3} and Table~\ref{tab:mtbench_specdec_draft_5} compares different speculative decoding configurations with and without CopySpec, using Qwen2.5-32B-Instruct as the target model and Qwen2.5-7B-Instruct as the draft model. This analysis explores the impact of varying $\gamma$ values and draft token counts, demonstrating that the integration of CopySpec with speculative decoding consistently leads to enhanced performance. The results emphasize the adaptability of CopySpec across diverse operational settings.

These tables collectively validate the effectiveness of CopySpec in accelerating large language model inference while maintaining high output quality. The findings in this appendix complement those in Appendix~\ref{sec:redundant}, reinforcing the method’s utility across datasets and configurations.

\section{Extra Results on GSM8K}

This appendix provides an in-depth analysis of the CopySpec approach applied to self-correcting tasks and speculative decoding. The results demonstrate the effectiveness of CopySpec in improving token processing speed, leveraging context repetition, and enhancing self-correction efficiency without compromising model accuracy.

\begin{table*}[ht]
\centering
\resizebox{\textwidth}{!}{%
\begin{tabular}{l
                c c c c
                c c c c
                c c c c
               }
\toprule
\textbf{Variant} 
& \multicolumn{4}{c}{\textbf{Turn 1}} 
& \multicolumn{4}{c}{\textbf{Turn 2}}
& \multicolumn{4}{c}{\textbf{Turn 3}} \\
\cmidrule(lr){2-5} \cmidrule(lr){6-9} \cmidrule(lr){10-13}
& \textbf{\% Copied} & \textbf{Tokens/s} & \textbf{$\bm{\tau_1}$} & \textbf{$\bm{\tau_2}$}
& \textbf{\% Copied} & \textbf{Tokens/s} & \textbf{$\bm{\tau_1}$} & \textbf{$\bm{\tau_2}$}
& \textbf{\% Copied} & \textbf{Tokens/s} & \textbf{$\bm{\tau_1}$} & \textbf{$\bm{\tau_2}$} \\
\midrule

Base Model   & --      & 10.25$\pm0.01$         & -- & -- 
            & --      & 10.17$\pm0.01$         & -- & -- 
            & --      & 8.68$\pm0.01$         & -- & -- \\

CopySpec ($\gamma=3$)   & 5.76\%      & 10.13$\pm0.01$         & 0.58 & -- 
            & 44.17\%      & \textbf{15.72}$\pm0.01$         & 4.90 & -- 
            & 82.79\%      & \textbf{21.89}$\pm0.01$         & 7.67 & -- \\

CopySpec ($\gamma=5$)   & 1.01\%      & 9.91$\pm0.02$         & 0.72 & -- 
            & 40.67\%      & 14.79$\pm0.01$         & 6.96 & --
            & 82.78\%      & 21.39$\pm0.02$         & 8.70 & -- \\

Spec. Dec.
            & --       & 12.92$\pm0.02$         & -- & 3.77    
            & --       & 12.27$\pm0.01$         & -- & 3.36    
            & --       & 11.44$\pm0.01$         & -- & 4.30 \\

Spec. Dec. + Copy ($\gamma=3$)
            & 1.47\%       & 12.67$\pm0.02$         & 0.53 & 3.77    
            & 40.23\%       & 14.65$\pm0.02$         & 6.08 & 2.52    
            & 81.18\%       & 20.81$\pm0.01$         & 7.71 & 3.39 \\

Spec. Dec. + Copy ($\gamma=5$)
            & 0.30\%       & \textbf{12.99}$\pm0.01$         & 0.55 & 3.78    
            & 38.93\%       & 14.95$\pm0.01$         & 7.81 & 2.59    
            & 81.84\%       & 21.51$\pm0.02$         & 8.72 & 3.40 \\

\bottomrule
\end{tabular}
}
\caption{Performance comparison for self-correcting tasks when the draft model generates 5 tokens at a time. Qwen2.5-32B-Instruct is the target model, and Qwen2.5-7B-Instruct is the draft model. \(\tau_1\) refers to the average tokens accepted by CopySpec, and \(\tau_2\) refers to the average number of tokens accepted by the draft model. The accuracy of the model improves by from 92\% to 93\%. The average TPS is highest for Spec. Dec. + Copy($\gamma$ = 5) at 15.59 while CopySpec alone achieves 14.84 TPS on average.}
\label{tab:specdec_selfcorrection_5tokens}
\end{table*}

Table~\ref{tab:specdec_selfcorrection_5tokens} extends the analysis to speculative decoding scenarios, focusing on the performance of CopySpec combined with speculative decoding when the draft model drafts 5 tokens at a time for self-correcting tasks. The table highlights the impact of varying draft model outputs, where CopySpec, combined with speculative decoding (\(\gamma=5\)), achieves the best overall performance. Metrics such as TPS and \(\tau\) show consistent improvements, with the approach accepting a higher average number of tokens per attempt. This configuration effectively balances the benefits of speculative decoding with CopySpec's ability to handle token repetition efficiently.

\begin{table*}[ht]
\centering
\resizebox{\textwidth}{!}{%
\begin{tabular}{l l
                c c c c
                c c c
                c c c c
               }
\toprule
\textbf{Model} & \textbf{Variant} 
& \multicolumn{4}{c}{\textbf{Turn 1}} 
& \multicolumn{3}{c}{\textbf{Turn 2}}
& \multicolumn{4}{c}{\textbf{Turn 3}} \\
\cmidrule(lr){3-6} \cmidrule(lr){7-9} \cmidrule(lr){10-13}
\textit{(Instruct)} & 
& \textbf{\% Copied} & \textbf{Tokens/s} & \textbf{$\bm{\tau}$} & \textbf{Acc}
& \textbf{\% Copied} & \textbf{Tokens/s} & \textbf{$\bm{\tau}$} 
& \textbf{\% Copied} & \textbf{Tokens/s} & \textbf{$\bm{\tau}$} & \textbf{Acc} \\
\midrule

\multirow{2}{*}{\textbf{Qwen2.5-72B}}
& CopySpec   & 6.12\% & 4.71$\pm0.01$ & 0.63 & \multirow{2}{*}{94\%}
            & 47.49\% & \textbf{7.49}$\pm0.01$ & 4.35 
            & 88.68\% & \textbf{10.59}$\pm0.01$ & 7.94 & \multirow{2}{*}{\textbf{96\%}} \\
& Base Model & -- & \textbf{4.74}$\pm0.01$ & -- &  
            & -- & 4.76$\pm0.01$ & -- 
            & -- & 3.98$\pm0.01$ & -- &  \\
\midrule

\multirow{2}{*}{\textbf{Qwen2.5-32B}}
& CopySpec   & 5.76\% & 10.13$\pm0.01$ & 0.58 & \multirow{2}{*}{92\%} 
            & 44.17\% & \textbf{15.72}$\pm0.01$ & 4.90 
            & 82.78\% & \textbf{21.89}$\pm0.01$ & 7.67 & \multirow{2}{*}{\textbf{93\%}} \\
& Base Model & -- & \textbf{10.25}$\pm0.01$ & -- & 
            & -- & 10.17$\pm0.01$ & -- 
            & -- & 8.68$\pm0.01$ & -- &  \\
\midrule

\multirow{2}{*}{\textbf{Qwen2.5-7B}}
& CopySpec   & 9.36\% & \textbf{41.01$\pm0.44$} & 0.87 & \multirow{2}{*}{84\%}
            & 60.34\% & \textbf{75.34$\pm0.68$} & 5.65 
            & 84.23\% & \textbf{93.68$\pm0.26$} & 7.35 & \multirow{2}{*}{\textbf{85\%}} \\
& Base Model & -- & 40.29$\pm0.02$ & -- & 
            & -- & 39.67$\pm0.05$ & -- 
            & -- & 35.63$\pm0.01$ & -- &  \\

\bottomrule
\end{tabular}
}
\caption{Performance comparison on the GSM8K dataset for self-correcting tasks across three turns, using CopySpec and the base model with Qwen2.5-Instruct variants. The table highlights significant improvements in tokens-per-second (TPS), percentage of tokens copied, and the number of tokens successfully copied (\(\tau\)) per attempt when attempting to copy 10 tokens, with \(\gamma=3\). These results demonstrate the effectiveness of CopySpec in leveraging increased context size and refining self-correction efficiency without compromising accuracy.}
\label{tab:selfcorrection}
\end{table*}

Table~\ref{tab:selfcorrection} compares the performance of CopySpec and baseline models across three turns using the GSM8K dataset for self-correcting tasks. The metrics include tokens-per-second (TPS), the percentage of tokens copied, and the number of tokens successfully copied (\(\tau\)) per attempt. CopySpec consistently achieves significant improvements, particularly in the second and third turns, where a larger context size enables better utilization of repetitive patterns. Notable gains are observed in TPS, with improvements exceeding 2× in some configurations, and the percentage of copied tokens highlights CopySpec's efficiency in refining self-corrections.

These results underscore the versatility of CopySpec in enhancing computational efficiency and self-correction capabilities across multiple scenarios. The combination of CopySpec with speculative decoding demonstrates its adaptability to diverse operational settings, paving the way for faster and more accurate large language model inference in tasks requiring iterative refinement.

\section{MT-Redundant Dataset Examples}
\label{sec:redundant_dataset}

This appendix provides one illustrative example from each of the eight categories in our new \textit{MT-Redundant} Dataset. \textit{MT-Redundant} builds upon MT-Bench by modifying the second turn of each conversation into a request for variations or adjustments of the first turn’s response, thus emulating real-world scenarios in which users seek revisions to previous outputs. Specifically, we replace the original second-turn prompt in MT-Bench \cite{zheng2023judgingllmasajudgemtbenchchatbot} with one that instructs the model to revisit and refine its previous answer. All assistant responses in this appendix are generated using Qwen2.5-72B-Instruct.

% 1. Writing (80–90)
\begin{figure*}[t]
    \includegraphics[width=\textwidth]{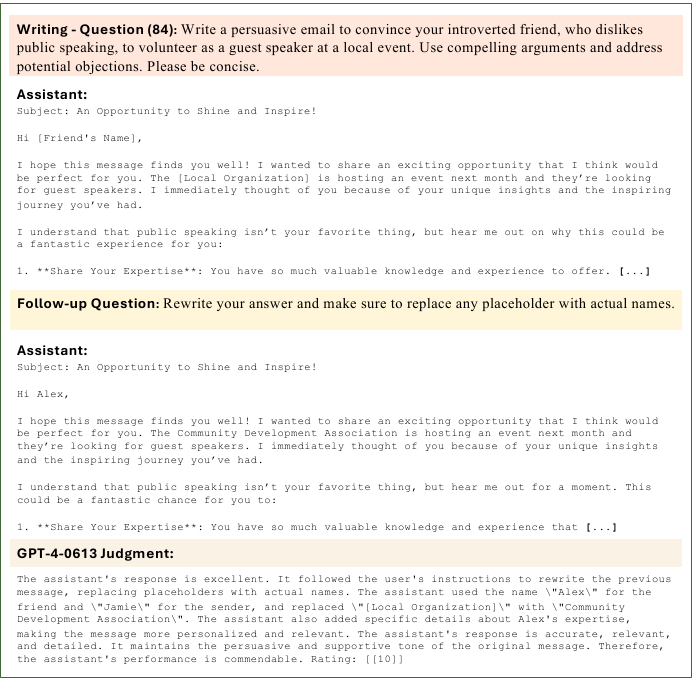}
    \caption{Examples from the \textbf{Writing} category (Slides 81--90). 
    This category focuses on creative and formal writing tasks, such as rephrasing, summarizing, or generating alternative drafts. 
    The second turn typically requests modifications or alternate versions of the initial written piece. "[...]" represents truncated output that didn't fit in the image.}
    \label{fig:writing}
\end{figure*}

% 2. Roleplay (91–100)
\begin{figure*}[t]
    \centering
    \includegraphics[width=\textwidth]{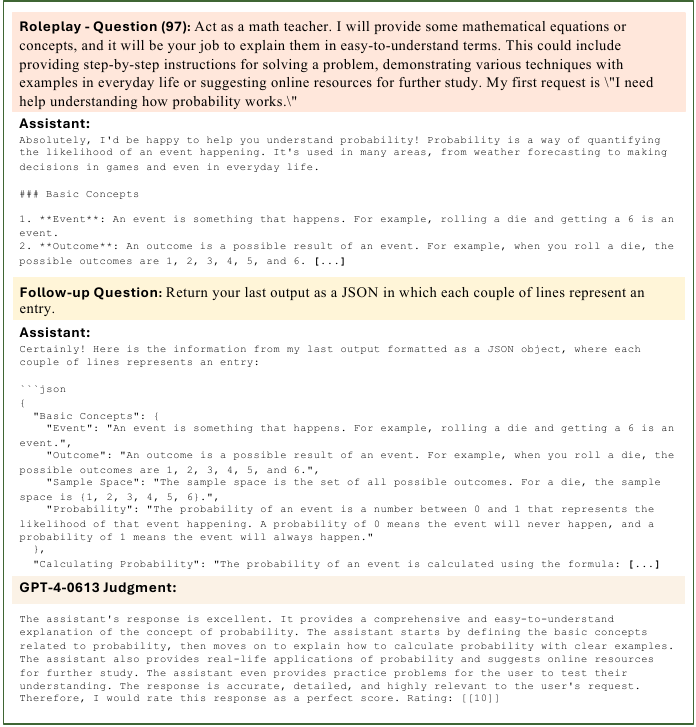}
    \caption{Examples from the \textbf{Roleplay} category (Slides 91--100). 
    Tasks in this category simulate real-world or imaginative scenarios, requiring the model to adjust its responses based on dynamic user requests and context shifts. "[...]" represents truncated output that didn't fit in the image.}
    \label{fig:roleplay}
\end{figure*}

% 3. Reasoning (101–110)
\begin{figure*}[t]
    \centering
    \includegraphics[width=\textwidth]{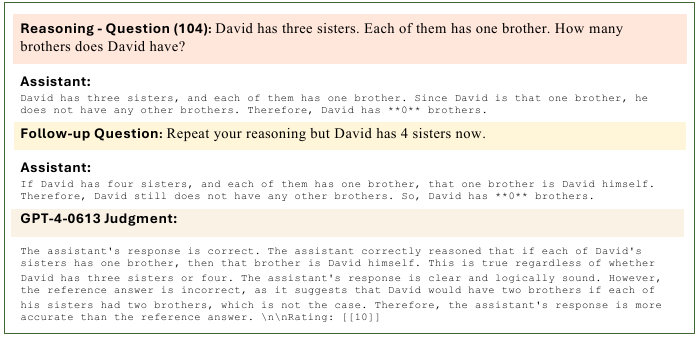}
    \caption{Examples from the \textbf{Reasoning} category (Slides 101--110). 
    This category evaluates logical and analytical thinking, with prompts asking models to refine prior explanations or provide additional clarifications in the second turn.}
    \label{fig:reasoning}
\end{figure*}

% 4. Math (111–120)
\begin{figure*}[t]
    \centering
    \includegraphics[width=\textwidth]{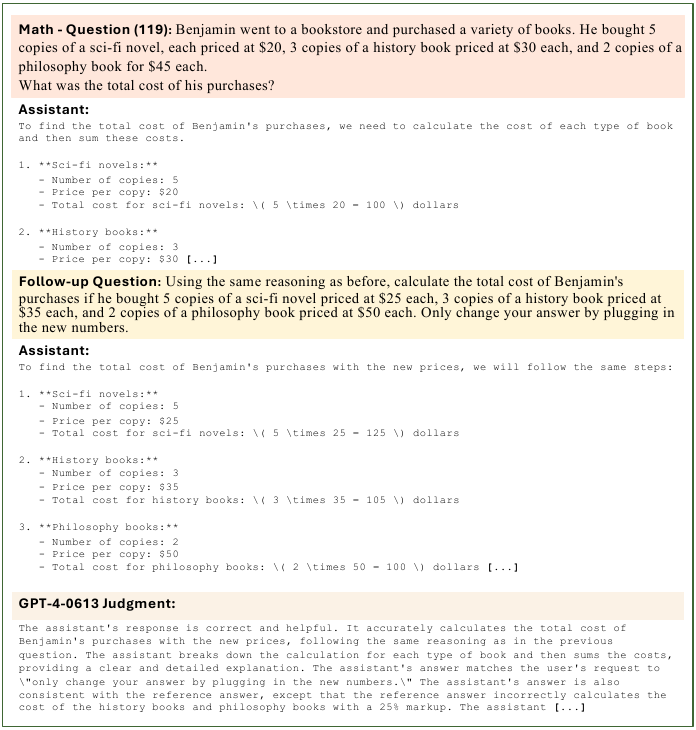}
    \caption{Examples from the \textbf{Math} category (Slides 111--120). 
    This category challenges the model to revise or elaborate mathematical solutions, often clarifying steps or offering alternative solution paths when asked. "[...]" represents truncated output that didn't fit in the image.}
    \label{fig:math}
\end{figure*}

% 5. Coding (121–130)
\begin{figure*}[t]
    \centering
    \includegraphics[width=\textwidth]{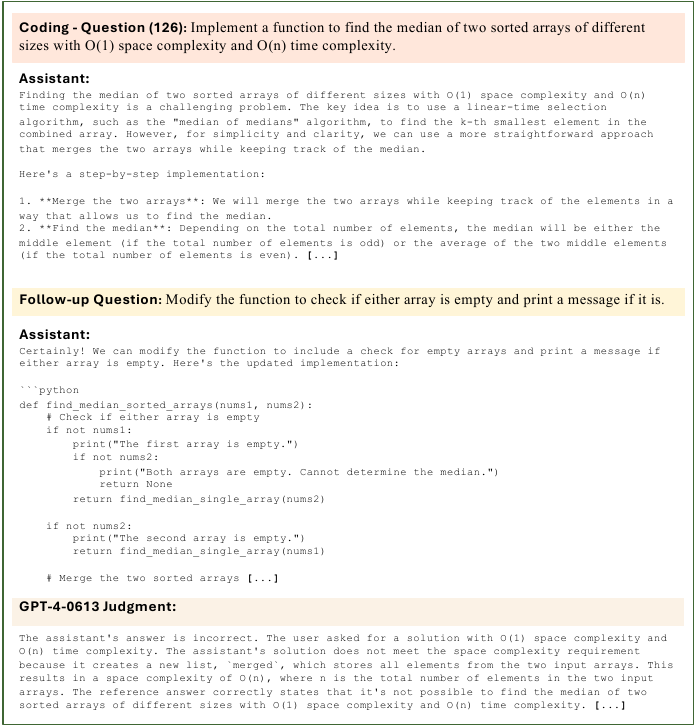}
    \caption{Examples from the \textbf{Coding} category (Slides 121--130). 
    This category covers programming-related tasks such as debugging, refactoring, or implementing variants of a provided code snippet in response to a user’s request. "[...]" represents truncated output that didn't fit in the image.}
    \label{fig:coding}
\end{figure*}

% 6. Extraction (131–140)
\begin{figure*}[t]
    \centering
    \includegraphics[width=\textwidth]{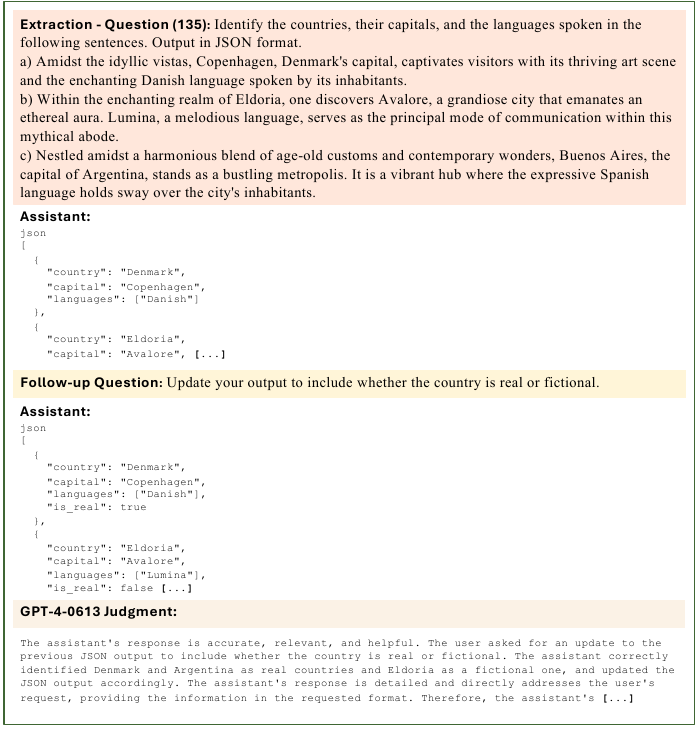}
    \caption{Examples from the \textbf{Extraction} category (Slides 131--140). 
    This category focuses on pulling specific information from the model’s previous response or restructuring it (e.g., lists, bullet points) according to user specifications. "[...]" represents truncated output that didn't fit in the image.}
    \label{fig:extraction}
\end{figure*}

% 7. STEM (141–150)
\begin{figure*}[t]
    \centering
    \includegraphics[width=\textwidth]{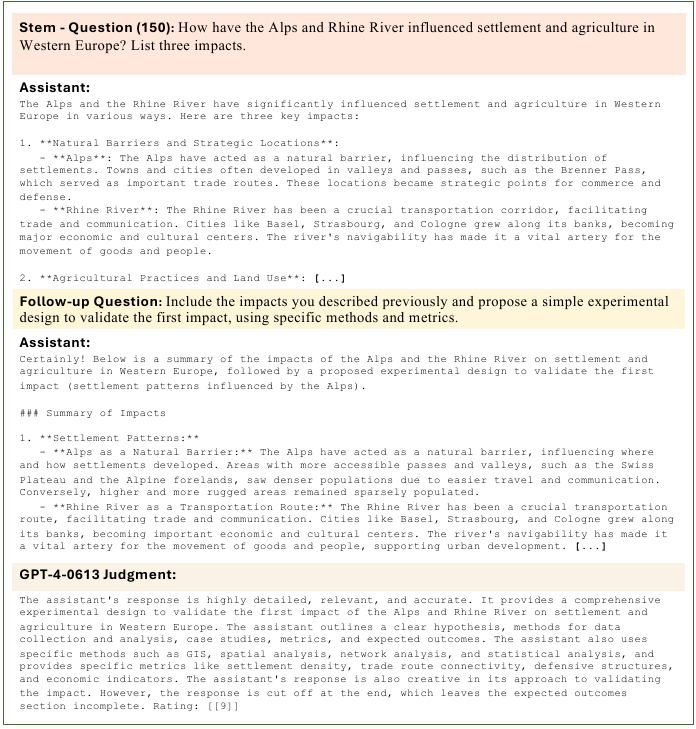}
    \caption{Examples from the \textbf{STEM} category (Slides 141--150). 
    This category addresses a variety of scientific and technical topics, requiring models to adapt or refine explanations, data, or methodologies in the second turn. "[...]" represents truncated output that didn't fit in the image.}
    \label{fig:stem}
\end{figure*}

% 8. Humanities (151–160)
\begin{figure*}[t]
    \centering
    \includegraphics[width=\textwidth]{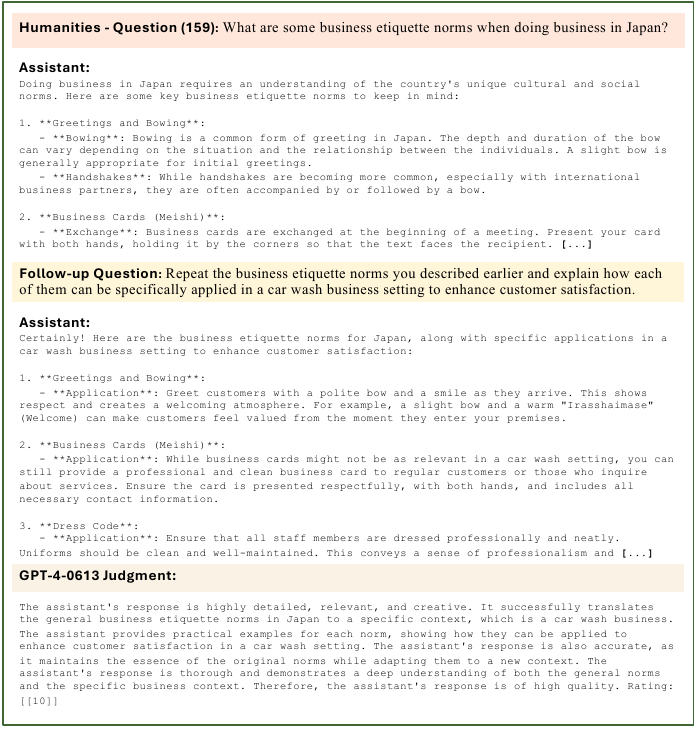}
    \caption{Examples from the \textbf{Humanities} category (Slides 151--160). 
    This category includes topics like literary analysis, historical context, or philosophical discussion, with the second turn often requesting deeper insight or alternate perspectives. "[...]" represents truncated output that didn't fit in the image.}
    \label{fig:humanities}
\end{figure*}

\section{Prompts Used}
\label{sec:prompts}

\subsection{Example of Self-Correction on GSM8K}
\label{sec:self-correcting}

This appendix presents an example of self-correction in code generation on the GSM8K dataset. Using Qwen2.5-72B-Instruct, we generate an initial solution and apply multi-round prompting to iteratively refine and correct the generated code.

To ensure direct answer generation, we prompt the model to explicitly print the computed result, reducing intermediate ambiguities and improving overall accuracy.

\begin{figure*}[t]
\centering
\includegraphics{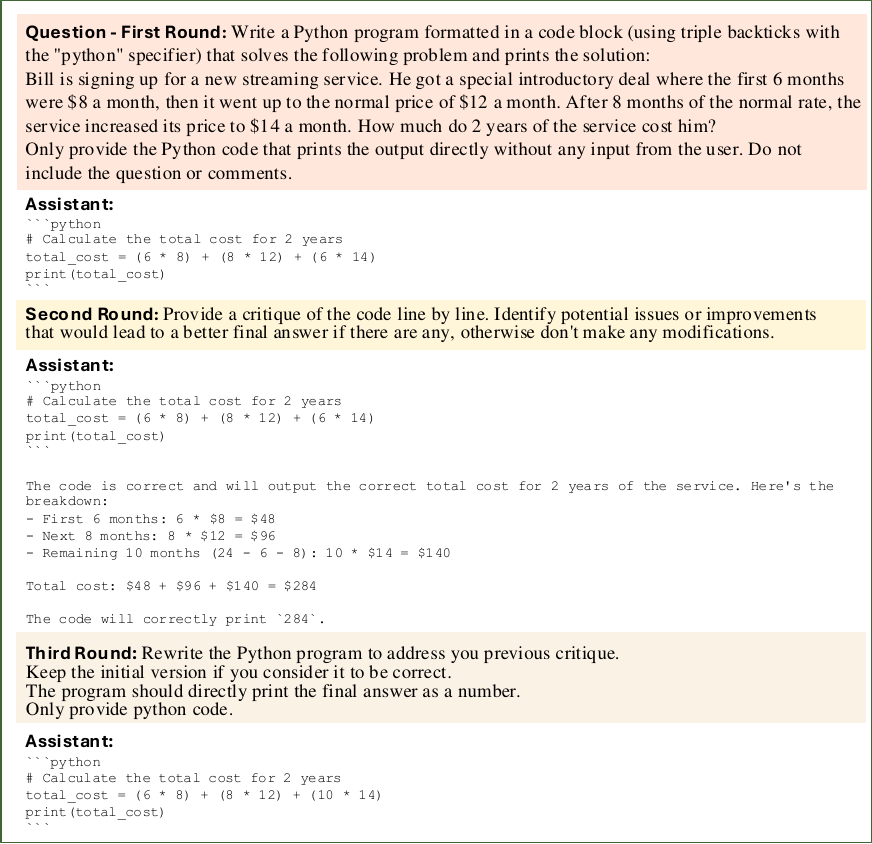}
\caption{An example of self-correction in code generation on the GSM8K dataset using Qwen2.5-72B-Instruct, showcasing iterative refinement to improve accuracy.}
%\label{fig:humanities}
\end{figure*}

\subsection{Example of Extractive Summarization}
\label{sec:extractive_summary}

This appendix provides an example of extractive summarization, where key sentences are selected directly from the original text to form a concise summary. The example, generated using Qwen2.5-72B-Instruct, demonstrates how to extract the most relevant information while preserving the original wording. Notably, the Qwen models show an interesting trend on the CNN/DM dataset, where larger models produce more extractive summaries that achieve slightly lower ROUGE-L scores.

\begin{figure*}[t]
\centering
\includegraphics[width=\textwidth]{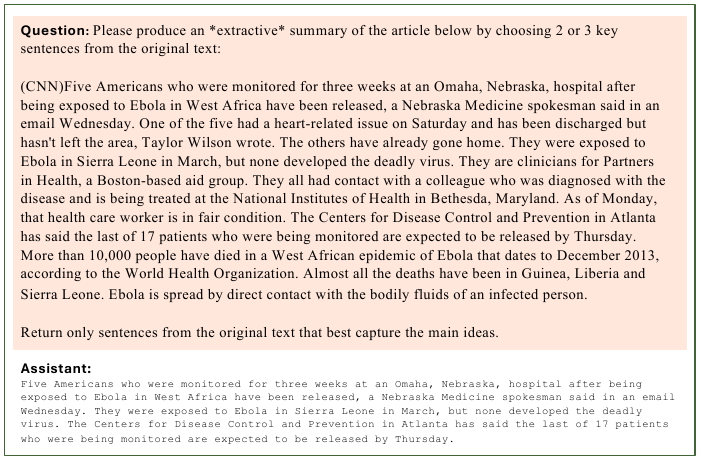}
\caption{An example of self-correction in code generation on CNN/DM using Qwen2.5-72B-Instruct, demonstrating abstractive summarization.}
%\label{fig:humanities}
\end{figure*}

\subsection{Code Generation on HumanEval}
\label{sec:code}

This section presents an example of code generation using Qwen2.5-72B-Instruct on the HumanEval dataset. The model generates an initial code implementation based on a given problem description and produces a self-contained Python script that correctly solves the task. The input consists of a problem description specifying the function signature, expected behavior, and an example test case. The generated solution includes function definitions, type hints, and example test cases to ensure correctness.

\begin{figure*}[t]
\centering
\includegraphics[width=\textwidth]{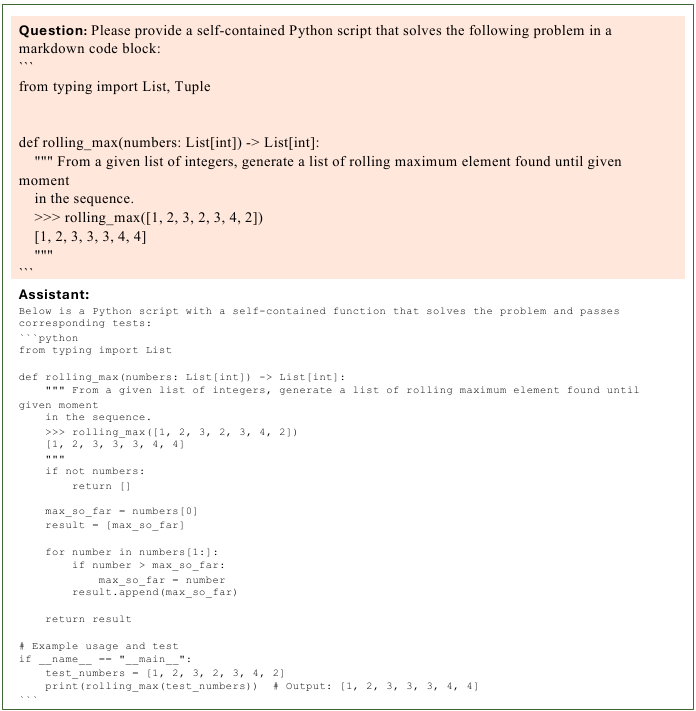}
\caption{Example of code generation on the HumanEval dataset using Qwen2.5-72B-Instruct, demonstrating the model's ability to produce a self-contained Python solution with function definitions, type hints, and test cases.}
\label{fig:humaneval}
\end{figure*}

\end{document}